\documentclass[lettersize,journal]{IEEEtran}
\usepackage{amsmath,amsfonts,amssymb}
\usepackage{algorithmic}
\usepackage{algorithm}
\usepackage{xcolor}
\usepackage{array}
\usepackage{lmodern} 
\usepackage{booktabs}
\usepackage{textcomp}
\usepackage{stfloats}
\usepackage{url}
\usepackage{verbatim}
\usepackage{graphicx}
\usepackage{booktabs}
\usepackage{dcolumn} 
\usepackage{cite}
\usepackage{subcaption}\usepackage{graphicx}
\usepackage{booktabs}
\usepackage{multirow}
\usepackage[table]{xcolor}
\usepackage{booktabs}
\usepackage{multirow}
\usepackage{makecell}
\usepackage{array}
 
\newcommand{\meanstd}[2]{$#1 \pm #2$}
\newcommand{\meanstdbold}[2]{$\mathbf{#1} \pm \mathbf{#2}$}

\begin{document}

\title{Spatial-Frequency-Aware Implicit Neural Representation of Multidimensional Signals via MLP-KAN Fusion}

\author{
Wen Yan, Ligen Shi, Jun Qiu, Haimiao Zhang,\IEEEmembership{Member, IEEE}, Lina Wu, \IEEEmembership{Member, IEEE} and Chang Liu*, \IEEEmembership{Member, IEEE}%
\thanks{\textit{Wen Yan and Ligen Shi contributed equally to this work.} \textit{Corresponding author: Chang Liu}.} %
\thanks{Wen Yan, Jun Qiu, Haimiao Zhang, Lina Wu and Chang Liu are with the Institute of Computational Imaging, Beijing Information Science and Technology University, Beijing 102206, China (e-mail: 2023020324@bistu.edu.cn; qiu.jun.cn@ieee.org; hmzhang@bistu.edu.cn; wulina@bistu.edu.cn; liu.chang.cn@ieee.org).\par
Ligen Shi is with the College of Computer Science (College of Software), Inner Mongolia University, Hohhot 010021, China, and also with the Research Center for Spatiotemporal Intelligence, Inner Mongolia University, Hohhot 010021, China (e-mail: ligenshi0826@gmail.com).}
}



\maketitle

\begin{abstract}
Implicit Neural Representations (INRs) have emerged as a compelling paradigm for modeling multidimensional signals by mapping continuous coordinates to signal values. However, Multi-Layer Perceptrons (MLP)-based INRs inherently suffer from spectral bias, which favors low-frequency components and suppresses the reconstruction of essential high-frequency details. While existing techniques, such as Fourier feature mappings, mitigate this issue, they often rely on sensitive manual tuning and are prone to spectral artifacts. In this paper, we propose a spatial-frequency-aware INR framework that combines an MLP branch with a Kolmogorov-Arnold network (KAN) branch for complementary frequency-oriented modeling. The MLP branch provides a low-frequency-oriented representation of smooth structures, whereas the KAN branch complements localized variations and fine details. To coordinate the two branches, we integrate the discrete wavelet transform (DWT) and inverse discrete wavelet transform (IDWT) into the output fusion stage. The outputs of the two branches are decomposed into wavelet coefficients, and the corresponding coefficients are additively fused before inverse wavelet reconstruction. A wavelet-domain band-separation regularization further penalizes high-frequency responses in the MLP branch and low-frequency responses in the KAN branch, thereby encouraging complementary frequency-oriented behavior. Experiments on 1D signals, 2D images, 3D volumes and signed distance functions, videos, and 4D light-fields demonstrate the applicability of the proposed representation across the evaluated signal modalities. Results demonstrate improved reconstruction fidelity across the evaluated signal modalities.

\end{abstract}


\section{Introduction}

\IEEEPARstart{I}{mplicit} Neural Representations (INRs) have emerged as a powerful paradigm for signal representation, replacing discrete grid-based tensors with coordinate-based neural functions~\cite{xie2022neural,essakine2025where}. By parameterizing a signal as a mapping from spatial, temporal, or other coordinates to its corresponding values, INR enables continuous-coordinate querying, compact parameterization, and differentiable signal modeling~\cite{luo2025continuous}. These properties have led to substantial advances in 3D surface modeling~\cite{park2019deepsdf,yang2025monge}, novel-view synthesis~\cite{mildenhall2021nerf,barron2021mip}, shape modeling~\cite{wu2025generalized}, and video representation~\cite{gao2025pnvc,zhu2025implicit}. Moreover, INRs have shown considerable potential for multidimensional signal reconstruction and inverse problems~\cite{zhang2026coarse}, including light-field (LF) representation and reconstruction~\cite{Liu:26,cong2026learning}, disparity estimation~\cite{shi2025iterative}, infrared and visible image fusion~\cite{11683540}, computed tomography reconstruction~\cite{zhou2025spatiotemporal, shen2025incomplete}, and ring artifact removal~\cite{shi2025ring}. These studies demonstrate the flexibility of INRs for representing and reconstructing signals with different dimensions and modalities.

\IEEEpubidadjcol

Despite their advantages, MLP-based INRs inherently exhibit spectral bias~\cite{rahaman2019spectral, bietti2019inductive}, whereby low-frequency components are learned preferentially, while high-frequency variations are more difficult to capture. Existing methods primarily attempt to overcome this behavior through coordinate encoding, activation-function design, or hybrid explicit--implicit representations. Fourier feature mappings~\cite{tancik2020fourier} embed coordinates into a higher-frequency space, but their performance depends on the selected encoding frequencies and dimensions. SIREN~\cite{sitzmann2020implicit} employs periodic activation functions, but requires carefully designed initialization and network configurations. Hybrid explicit--implicit representations, such as InstantNGP~\cite{muller2022instant} and DINER~\cite{xie2023diner}, augment coordinate networks with trainable embeddings and introduce additional signal-dependent storage overhead. Although these approaches improve high-frequency representation, they primarily address spectral bias by modifying the input representation or the MLP-based coordinate network itself. Taking a different perspective on spectral bias, we retain and exploit the low-frequency-oriented behavior of MLP and investigate a heterogeneous representation in which different network structures exhibit complementary frequency-oriented modeling behaviors.

\begin{figure*}[!t]
    \centering
    \includegraphics[width=0.95\textwidth]{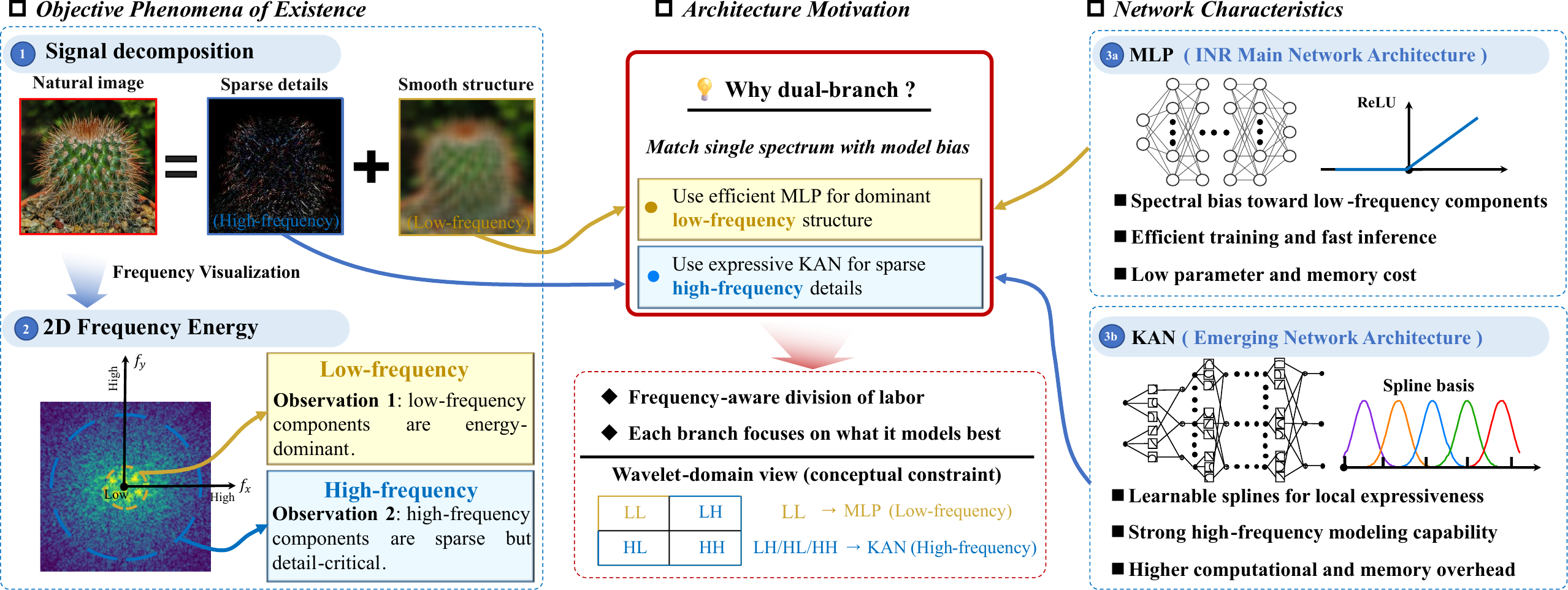}
    \caption{
    Motivation of the proposed spatial--frequency-aware MLP--KAN framework. The MLP branch provides a low-frequency-oriented representation of smooth structures, and the KAN branch provides complementary responses to localized variations and fine-scale details. DWT exposes the multiscale responses of both branches. Corresponding wavelet coefficients are combined through additive wavelet-domain fusion, and band-separation regularization encourages complementary frequency behavior between the two branches.}
    \label{fig:motivation}
\end{figure*}

As illustrated in Fig.~\ref{fig:motivation}, the proposed framework combines an MLP branch and a KAN branch to exploit their complementary representation characteristics. The MLP branch provides a low-frequency-oriented representation for smooth signal components, consistent with the spectral bias of conventional MLP. This characteristic is well matched to the spectral distribution of many natural signals, in which low-frequency components account for most of the signal energy and determine the global structure, whereas high-frequency components are relatively sparse but encode perceptually important edges, textures, and localized variations. In contrast to MLP, KAN~\cite{liu2025kan} parameterize network edges using learnable univariate functions, which are represented by spline basis functions in the standard spline-based implementation~\cite{unser2002b}, providing greater local flexibility for representing nonlinear variations. Recent analyses further indicate that KAN can exhibit a weaker low-frequency bias under appropriate configurations~\cite{wang2025expressiveness}. Accordingly, we retain the MLP branch for the dominant smooth structures and use the KAN branch as a complementary representation pathway with relatively stronger responses to localized variations and fine-scale details.  The two branches are not restricted to disjoint frequency ranges, and both contribute to the final reconstruction.

To coordinate the two representation pathways, we introduce wavelet-domain fusion and spectral regularization. Wavelet transforms provide localized multiresolution analysis of non-stationary signals. Based on this property, the outputs of the two branches are decomposed by DWT~\cite{zhang2025wavelet}. The corresponding wavelet coefficients generated by the MLP and KAN branches are then combined by element-wise addition, followed by IDWT reconstruction. This additive fusion allows both branches to contribute to each wavelet band and preserves the flexibility of the dual-branch representation.

Because wavelet decomposition alone does not guarantee complementary frequency responses between the two branches, we further introduce a band-separation loss that penalizes the high-frequency responses of the MLP branch and the low-frequency responses of the KAN branch. The resulting spectral constraint strengthens the low-frequency tendency of the MLP branch and increases the relative detail response of the KAN branch. The corresponding wavelet coefficients remain additively fused, so the regularization controls the frequency tendency of the branches without enforcing exclusive subband assignment. In this manner, the proposed framework alleviates the high-frequency fitting limitation of MLP-based INRs without discarding their advantageous low-frequency inductive bias, while preserving contributions from both branches to the final reconstruction. 

In summary, the main contributions of this work are as follows:

\begin{itemize}
    \item We propose a spatial--frequency-aware INR framework based on an MLP-KAN dual-branch architecture.
    The MLP branch provides a low-frequency-oriented representation of smooth components, and the KAN branch supplies complementary responses to localized and fine-scale variations. Their heterogeneous representation characteristics are jointly exploited for multidimensional signal fitting.

    \item 
    We introduce a wavelet-domain coordination mechanism based on additive coefficient fusion and branch-specific band-separation regularization. Corresponding DWT coefficients from the two branches are additively fused before inverse reconstruction, and the spectral regularization suppresses high-frequency responses in the MLP branch and low-frequency responses in the KAN branch to promote complementary frequency-oriented behavior.

    \item We systematically evaluate the proposed framework on diverse multidimensional signal representation tasks, including 1D signals, 2D images, 3D volumes and signed distance functions, spatiotemporal videos, and 4D LFs.
\end{itemize}

\section{Related Work}
This section reviews three lines of work most relevant to our method: implicit neural representations, Kolmogorov-Arnold networks, and wavelet-based multiresolution analysis.

\subsection{Implicit Neural Representation}
To improve high-frequency modeling, existing INRs modify coordinate encodings, activation functions, or network structures. Fourier features and Random Fourier Features~\cite{tancik2020fourier} map coordinates to higher-frequency spaces, while SIREN~\cite{sitzmann2020implicit}, GAUSS~\cite{ramasinghe2022beyond}, WIRE~\cite{saragadam2023wire}, and SINC~\cite{saratchandran2024sampling} use sinusoidal, Gaussian, Gabor-wavelet, or sinc activations. FINER~\cite{liu2024finer,zhu2026finer++} adapts activation frequencies, INCODE~\cite{kazerouni2024incode} modulates periodic activations, MFN~\cite{fathony2020multiplicative} employs multiplicative filters, and FR~\cite{shi2024improved} uses Fourier reparameterization. Recent multidimensional INRs further incorporate semantic structure~\cite{li2024superpixel} or learnable latent frequency spaces~\cite{zhou2025frequency}. Other recent approaches explicitly consider cross-frequency representation~\cite{yu2025cross} or joint frequency and locality information~\cite{ko2025flair}. Continuous tensor representations have also been developed to exploit structural priors for high-dimensional signal recovery~\cite{luo2024lowranktensor}. Our framework instead coordinates heterogeneous MLP/KAN representations through wavelet-domain fusion and branch-specific spectral regularization.

\subsection{Advances in Kolmogorov-Arnold Networks}
Inspired by the Kolmogorov-Arnold representation theorem, KAN~\cite{liu2025kan} replaces the fixed activations used in MLP with learnable univariate functions on network edges. WaveKAN~\cite{liu2026wavekan} combines adaptive wavelet decomposition with KAN for long-term time-series forecasting, using wavelet decomposition to separate low-frequency trends and high-frequency details before subsequent modeling and fusion. In addition, numerous variants based on KAN exist, such as GKAN~\cite{LiuXuLeiWangDuLv2026DGKAN}, RKAN~\cite{aghaei2026rkan}, KKAN~\cite{toscano2024kkans}, and FC-KAN~\cite{ta2024fc}. In this work, we adopt the standard spline-based KAN formulation to focus on the complementary representation behavior between heterogeneous MLP and KAN branches. The standard formulation provides a representative and relatively simple KAN architecture for evaluating this heterogeneous dual-branch design.

\subsection{Wavelet-based Multiresolution Analysis}
The DWT decomposes signals into approximation and detail subbands for multiresolution analysis~\cite{lucas2025multidimensional,11181153}. Wavelet and wavelet-frame representations have been widely studied in image restoration and related inverse problems~\cite{cai2012image}. Wavelet-based representations have also been incorporated into INRs
~\cite{saragadam2023wire}. Compared with global Fourier bases, wavelets provide localized spatial--frequency analysis, which is beneficial for representing non-stationary structures such as edges, textures, and local variations. Adaptive WPE~\cite{zhao2025adaptive} further incorporates wavelet information into coordinate encoding, whereas our framework applies DWT to branch outputs for coefficient fusion and band-separation regularization. However, applying DWT alone does not automatically lead to frequency specialization between different neural branches. Since DWT is a linear transform, directly decomposing branch outputs and recombining the corresponding subbands does not necessarily change the modeling behavior unless additional constraints are introduced. This observation motivates the use of wavelet-domain regularization to explicitly measure and constrain the frequency responses of different branches.

\section{Method}
\subsection{Architecture Overview}
\begin{figure} [!t]
\begin{center}
\begin{tabular}{c}
\includegraphics[width=0.5\textwidth]{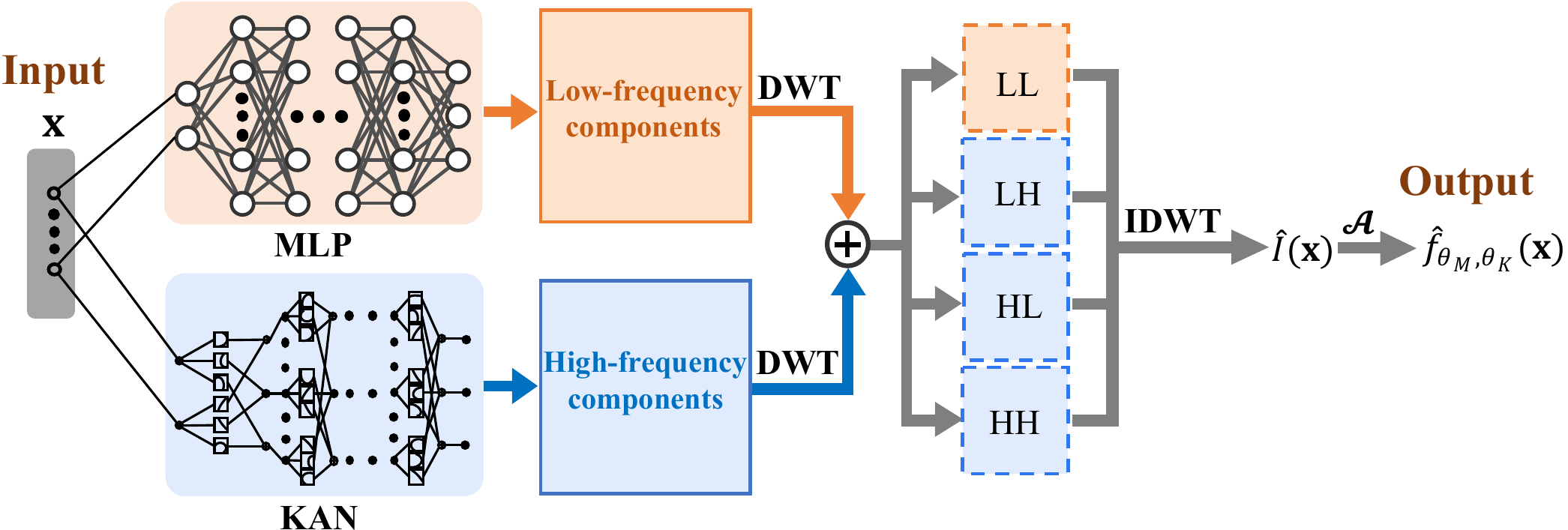}
\end{tabular}
\end{center}
\caption{Architecture of the proposed method. The input coordinates are processed by parallel MLP and KAN branches that provide complementary frequency-oriented representations. Their outputs are decomposed by DWT, corresponding wavelet coefficients are additively fused, and the fused coefficients are reconstructed by IDWT. A band-separation regularization encourages complementary frequency responses between the two branches.}
\label{fig-1}
\end{figure}

We propose a spatial-frequency-aware INR framework based on an
MLP-KAN dual-branch architecture with wavelet-domain fusion and regularization, as shown in Fig.~\ref{fig-1}.

The proposed framework maps continuous coordinates to signal values through two parallel branches. The MLP branch provides a low-frequency-oriented representation, whereas the KAN branch provides a complementary representation with relatively stronger responses to localized variations and fine-scale details. The outputs of the two branches are decomposed into wavelet coefficients by DWT, and the corresponding coefficients are combined through element-wise addition. The fused coefficients are subsequently reconstructed by IDWT.

Because additive fusion does not impose a hard frequency assignment, both branches can contribute to the reconstructed signal. We therefore introduce a wavelet-domain band-separation regularization to encourage complementary frequency behavior by suppressing high-frequency responses in the MLP branch and low-frequency responses in the KAN branch.

\subsection{Dual-Branch Spectral Decomposition}
Given an input coordinate $\mathbf{x}\in\mathbb{R}^{d_{\mathrm{in}}}$, 
the proposed framework employs two coordinate networks to parameterize complementary spectral representations. After wavelet-domain fusion and IDWT reconstruction, the final prediction is expressed as the sum of a low-frequency component and a high-frequency component:
\begin{equation}
\hat{\mathbf{I}}=\hat{\mathbf{I}}_{\mathrm{low}}+\hat{\mathbf{I}}_{\mathrm{high}},
\label{eq:low_high_decomposition}
\end{equation}
where $\hat{\mathbf{I}}_{\mathrm{low}}$ and $\hat{\mathbf{I}}_{\mathrm{high}}$ denote the approximation-band and detail-band components of the fused reconstruction, respectively. Under the band-separation constraint, these components are encouraged to be dominated by the MLP and KAN branches, respectively.

\subsubsection{MLP Branch for Global Structure}

The MLP branch parameterizes the smooth global structure of the signal. 
Let $\mathbf{h}_{0}=\mathbf{x}$. For an $L_M$-layer MLP, the hidden features are computed as
\begin{equation}
\mathbf{h}_{l}=\sigma\!\left(\mathbf{W}_{l}\mathbf{h}_{l-1}+\mathbf{b}_{l}\right),\quad l=1,\ldots,L_M-1,
\label{eq:mlp_hidden}
\end{equation}
where $\sigma(\cdot)$ denotes a fixed nonlinear activation function. 
The output of the MLP branch is defined by
\begin{equation}
f_{\theta_M}(\mathbf{x})=\mathbf{W}_{L_M}\mathbf{h}_{L_M-1}+\mathbf{b}_{L_M},
\label{eq:mlp_mapping}
\end{equation}
where $\theta_M=\{\mathbf{W}_{l},\mathbf{b}_{l}\}_{l=1}^{L_M}$ contains the learnable parameters of the MLP branch. The output $f_{\theta_M}(\mathbf{x})$ serves as a low-frequency-oriented representation and is encouraged to capture slowly varying global structures.

\subsubsection{KAN Branch for Localized Details}
The KAN branch parameterizes localized signal variations using learnable univariate functions placed on network edges. Let $\mathbf{z}_{0}=\mathbf{x}$. For the $\ell$-th KAN layer, the $q$-th output feature is computed as
\begin{equation}
z_{\ell+1,q}=\sum_{p=1}^{d_{\ell}}\phi_{\ell,q,p}\!\left(z_{\ell,p}\right),
\quad
q=1,\ldots,d_{\ell+1},
\label{eq:kan_layer}
\end{equation}
where $d_{\ell}$ is the feature dimension and $\phi_{\ell,q,p}(\cdot)$ is the learnable function on the edge from input $p$ to output $q$.

In the spline-based implementation, each edge function is parameterized as
\begin{equation}
\phi_{\ell,q,p}(t)=w_{\ell,q,p} b(t)+\sum_{r=1}^{R}c_{\ell,q,p,r} B_r(t),
\label{eq:spline_edge_function}
\end{equation}
where $b(\cdot)$ is the base function, $B_r(\cdot)$ is the $r$-th B-spline basis, and $w_{\ell,q,p}$ and $c_{\ell,q,p,r}$ are learnable coefficients. $B_r(\cdot)$ denotes the $r$-th B-spline basis function of degree $k$. The spline grid size controls the number of knot intervals used to construct the basis functions. The degree $k$ and grid size reported in the experiments therefore determine the local spline order and resolution of each KAN edge function, respectively.

After recursively applying Eq.~\eqref{eq:kan_layer} over $L_K$ layers, the output of the KAN branch is defined as
\begin{equation}
g_{\theta_K}(\mathbf{x})=\mathbf{z}_{L_K}, 
\label{eq:kan_mapping}
\end{equation}
where $\theta_K$ collects all learnable parameters of the KAN branch. The output $g_{\theta_K}(\mathbf{x})$ serves as a complementary representation with relatively stronger responses to localized rapid variations, edges, textures, and fine-scale details.

\subsection{Wavelet-Domain Fusion and Reconstruction}
Let $\mathcal{W}_{\psi}(\cdot)$ and $\mathcal{W}_{\psi}^{-1}(\cdot)$ denote DWT and IDWT with wavelet basis $\psi$. We use a $J$-level 1D Haar (db1) transform along sampled coordinate-axis lines; Eqs.~\eqref{eq:dwt_mlp}--\eqref{eq:high_component} use a one-level 2D formulation for clarity.

Given a discrete coordinate grid $\mathcal{X}=\{\mathbf{x}_{n}\}_{n=1}^{N}$, the two coordinate networks are evaluated to obtain
\begin{equation}
\mathbf{F}_{M}=f_{\theta_M}(\mathcal{X}),
\qquad
\mathbf{F}_{K}=g_{\theta_K}(\mathcal{X}),
\label{eq:branch_grid_outputs}
\end{equation}
where $\mathbf{F}_{M}$ and $\mathbf{F}_{K}$ denote the signal tensors generated by the MLP and KAN branches, respectively.

The MLP branch output is decomposed as
\begin{equation}
\left\{Y_{LL}^{M},Y_{LH}^{M},Y_{HL}^{M},Y_{HH}^{M}\right\}=\mathcal{W}_{\psi}\!\left(\mathbf{F}_{M}\right),
\label{eq:dwt_mlp}
\end{equation}
and the KAN branch output is decomposed as
\begin{equation}
\left\{Y_{LL}^{K},Y_{LH}^{K},Y_{HL}^{K},Y_{HH}^{K}\right\}=\mathcal{W}_{\psi}\!\left(\mathbf{F}_{K}\right).
\label{eq:dwt_kan}
\end{equation}
Here, $Y_{LL}^{(\cdot)}$ denotes the low-frequency approximation subband,
whereas $Y_{LH}^{(\cdot)}$, $Y_{HL}^{(\cdot)}$, and $Y_{HH}^{(\cdot)}$ denote the three directional high-frequency detail subbands.

\begin{figure}[!t]
    \centering
    \includegraphics[width=0.45\textwidth]{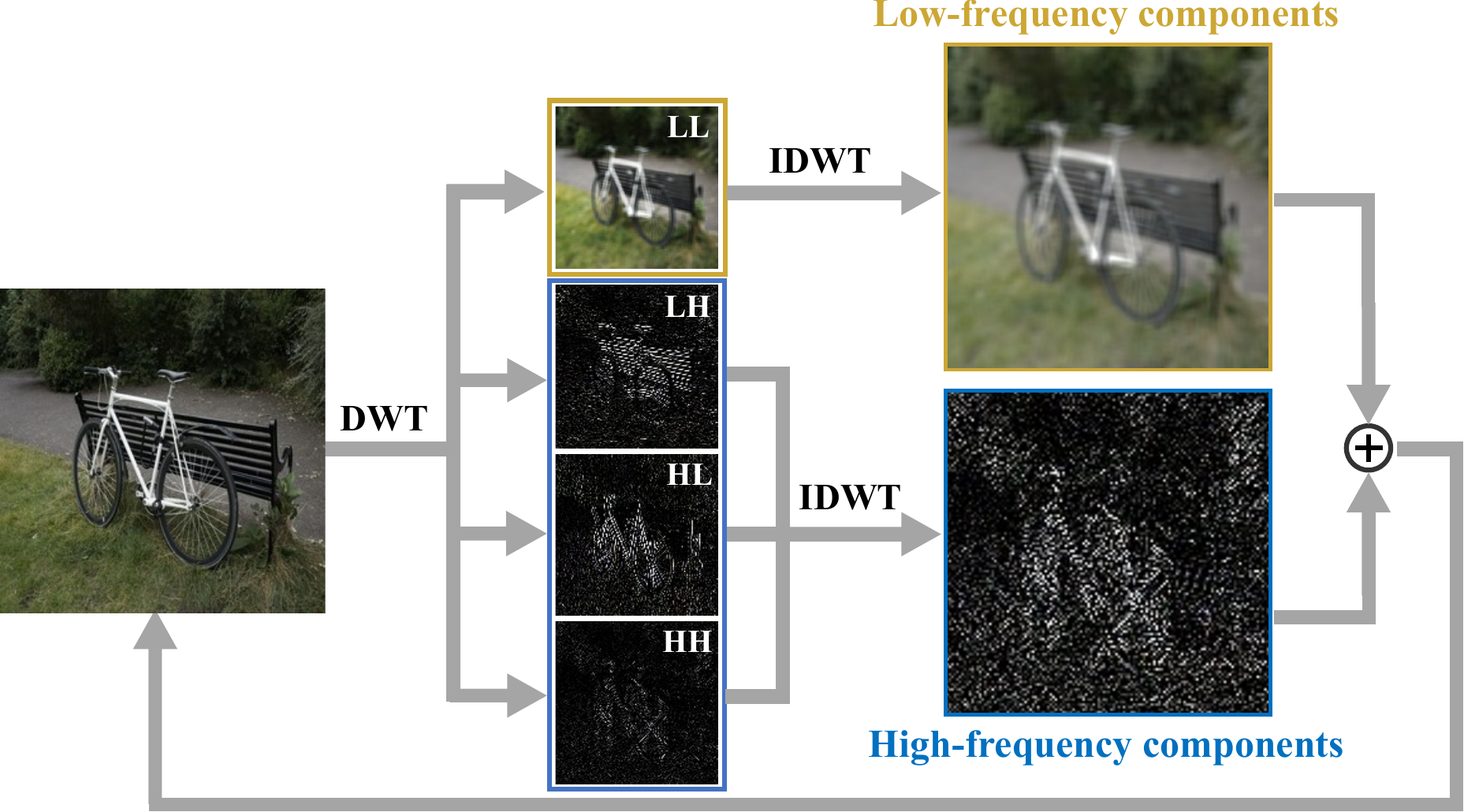}
    \caption{Schematic illustration of wavelet-domain decomposition and reconstruction. For a 2D signal, DWT decomposes the representation into the LL, LH, HL, and HH subbands. Corresponding MLP and KAN wavelet coefficients are additively fused before IDWT reconstruction, while band-separation regularization encourages complementary frequency-oriented responses.}
    \label{fig-2}
\end{figure}

The corresponding coefficients are additively fused as
\(\widetilde{Y}_{b}^{F}=Y_{b}^{M}+Y_{b}^{K}\),
\(b\in\{LL,LH,HL,HH\}\), allowing both branches to contribute to each band.
Band-separation regularization encourages the branch responses toward the following idealized limiting state:

\begin{equation}
\left\{
\begin{aligned}
\widetilde{Y}_{LL}^{F} &= Y_{LL}^{M},\\
\widetilde{Y}_{LH}^{F} &= Y_{LH}^{K},\\
\widetilde{Y}_{HL}^{F} &= Y_{HL}^{K},\\
\widetilde{Y}_{HH}^{F} &= Y_{HH}^{K}.
\end{aligned}
\right.
\label{eq:frequency_assignment}
\end{equation}

Eq.~\eqref{eq:frequency_assignment} represents the limiting case in which the high-frequency responses of the MLP branch and the low-frequency response of the KAN branch are fully suppressed.

The final prediction is reconstructed from the fused subbands through IDWT:
\begin{equation}
\begin{aligned}
\hat{\mathbf{I}}
&=\hat{f}_{\theta_M,\theta_K}(\mathcal{X})\\
&=\mathcal{W}_{\psi}^{-1}
\left(
\widetilde{Y}_{LL}^{F},
\widetilde{Y}_{LH}^{F},
\widetilde{Y}_{HL}^{F},
\widetilde{Y}_{HH}^{F}
\right).
\end{aligned}
\label{eq:idwt_reconstruction}
\end{equation}

To characterize the approximation and detail components of the actual fused reconstruction, we define:
\begin{equation}
\hat{\mathbf{I}}_{\mathrm{low}}
=
\mathcal{W}_{\psi}^{-1}
\left(
\widetilde{Y}_{LL}^{F},
\mathbf{0},
\mathbf{0},
\mathbf{0}
\right),
\label{eq:low_component}
\end{equation}
and
\begin{equation}
\hat{\mathbf{I}}_{\mathrm{high}}
=
\mathcal{W}_{\psi}^{-1}
\left(
\mathbf{0},
\widetilde{Y}_{LH}^{F},
\widetilde{Y}_{HL}^{F},
\widetilde{Y}_{HH}^{F}
\right).
\label{eq:high_component}
\end{equation}
Here, $\mathbf{0}$ is introduced only to isolate the approximation or detail component during inverse reconstruction; it does not imply that the corresponding coefficients of the actual fused representation are zero.

For multidimensional signals, we sample complete coordinate-axis lines and apply a $J$-level 1D DWT/IDWT along each line rather than constructing a full $D$-dimensional wavelet tensor. Approximation coefficients are recursively decomposed and all detail coefficients are retained. For axis $d$, the fused coefficients are
$\widetilde{Y}_{l}^{F,d}=Y_{l}^{M,d}+Y_{l}^{K,d}$ and
$\widetilde{Y}_{h,j}^{F,d}=Y_{h,j}^{M,d}+Y_{h,j}^{K,d}$,
$j=1,\ldots,J$. The same fusion and band-separation regularization are applied along every sampled axis, including horizontal and vertical lines in 2D and their extensions to higher dimensions.

\subsection{Objective Function and Spectral Constraints}
For direct signal representation, reconstruction fidelity is measured by
\begin{equation}
\mathcal{L}_{\mathrm{recon}}=\frac{1}{N}\left\|\hat{\mathbf{I}}-\mathbf{I}\right\|_{F}^{2},
\label{eq:reconstruction_loss}
\end{equation}
where $\hat{\mathbf{I}}$ and $\mathbf{I}$ denote the reconstructed and ground-truth signals, respectively, $N$ denotes the total number of scalar elements in the signal tensor, and $\|\cdot\|_{F}$ denotes the Frobenius norm.

Because additive fusion alone does not induce complementary frequency-oriented responses between the two branches, the two branches may retain responses outside their intended frequency tendencies. We therefore introduce a wavelet-domain band-separation loss to encourage the idealized limiting state in Eq.~\eqref{eq:frequency_assignment} by suppressing the high-frequency responses of the MLP branch and the low-frequency response of the KAN branch. For consistency with the one-level 2D formulation in Eqs.~\eqref{eq:dwt_mlp}--\eqref{eq:high_component}, let $\mathcal{B}_{H}$ denote the set of high-frequency detail subbands. The band-separation loss is defined as
\begin{equation}
\mathcal{L}_{\mathrm{band}}=\alpha\sum_{b\in\mathcal{B}_{H}}
\frac{1}{N_b}
\left\|
Y_{b}^{M}
\right\|_{F}^{2}
+
(1-\alpha)
\frac{1}{N_{LL}}
\left\|
Y_{LL}^{K}
\right\|_{F}^{2},
\label{eq:band_separation_loss}
\end{equation}
where $Y_b^{M}$ denotes a high-frequency detail subband generated by the MLP branch, $Y_{LL}^{K}$ denotes the low-frequency approximation subband generated by the KAN branch, and $N_b$ and $N_{LL}$ are the corresponding numbers of wavelet coefficients. The parameter $\alpha\in[0,1]$ controls the relative weighting between the aggregated MLP high-frequency penalty and the KAN low-frequency penalty, and $\mathcal{B}_{H}=\{LH,HL,HH\}$ for the one-level 2D formulation. For the actual $J$-level line-wise implementation described above, the same penalty is extended to the corresponding approximation and detail coefficients along each sampled axis. By penalizing the MLP high-frequency responses and the KAN low-frequency response, $\mathcal{L}_{\mathrm{band}}$ promotes complementary frequency responses between the two branches without imposing hard subband assignment.

The overall objective for direct signal representation is formulated as
\begin{equation}
\mathcal{L}
=
\mathcal{L}_{\mathrm{recon}}
+
\lambda_{\mathrm{band}}
\mathcal{L}_{\mathrm{band}},
\label{eq:overall_direct_objective}
\end{equation}
where $\lambda_{\mathrm{band}}\geq 0$ controls the strength of the spectral constraint. This objective encourages a stronger low-frequency tendency in the MLP branch and relatively stronger detail responses in the KAN branch while preserving additive coefficient fusion.

\subsection{Inverse Problem Formulation}
The proposed representation can also be incorporated into inverse problems. Let $\mathcal{A}$ denote a known forward operator that maps a latent signal $\mathbf{I}$ to the observation domain. The measurement model is expressed as
\begin{equation}
\mathbf{y}
=
\mathcal{A}(\mathbf{I})
+
\boldsymbol{\epsilon},
\label{eq:measurement_model}
\end{equation}
where $\mathbf{y}$ denotes the observed data and $\boldsymbol{\epsilon}$ represents measurement noise.

Given the reconstructed signal $\hat{\mathbf{I}}$ produced by the proposed MLP-KAN representation, the corresponding data-fidelity loss is defined as
\begin{equation}
\mathcal{L}_{\mathrm{data}}
=
\frac{1}{N_y}
\left\|
\mathcal{A}(\hat{\mathbf{I}})
-
\mathbf{y}
\right\|_{2}^{2},
\label{eq:data_fidelity_loss}
\end{equation}
where $N_y$ is the number of observed measurements. Direct signal representation is a special case of Eq.~\eqref{eq:data_fidelity_loss}, in which $\mathcal{A}$ is the identity operator and $\mathbf{y}=\mathbf{I}$.

For inverse reconstruction, the network parameters are estimated by
\begin{equation}
\left(
\theta_M^{*},
\theta_K^{*}
\right)
=
\underset{\theta_M,\theta_K}{\arg\min}
\left[
\mathcal{L}_{\mathrm{data}}
+
\lambda_{\mathrm{band}}
\mathcal{L}_{\mathrm{band}}
+
\lambda_{\mathrm{TV}}
\mathcal{R}_{\mathrm{TV}}(\hat{\mathbf{I}})
\right],
\label{eq:inverse_objective}
\end{equation}
where $\lambda_{\mathrm{band}}\geq0$ and
$\lambda_{\mathrm{TV}}\geq0$ control the spectral constraint and spatial regularization, respectively. The reconstructed latent signal is subsequently obtained by evaluating the
proposed representation using the optimized network parameters:
\begin{equation}
\hat{\mathbf{I}}^{*}
=
\hat{f}_{\theta_M^{*},\theta_K^{*}}(\mathcal{X}).
\label{eq:optimal_reconstruction}
\end{equation}

For regularly sampled multidimensional signals, the anisotropic total variation regularizer is defined as
\begin{equation}
\mathcal{R}_{\mathrm{TV}}(\hat{\mathbf{I}})
=
\sum_{d=1}^{D}
\left\|
\mathbf{D}_{d}\hat{\mathbf{I}}
\right\|_{1},
\label{eq:discrete_tv}
\end{equation}
where $D$ is the signal dimensionality and $\mathbf{D}_{d}$ denotes the discrete difference operator along the $d$-th dimension. The $\ell_1$-norm promotes sparsity of the signal gradients, thereby suppressing noise while preserving sharp structural transitions.

\section{Experiments}

\subsection{Experimental Setup}
\label{sec:experimental_setup}
We evaluate per-instance fitting on 1D square waves and audio, 2D images, 3D volumes and SDFs, videos, and 4D LFs. These tasks encompass piecewise-constant signals with abrupt transitions, oscillatory temporal signals, natural images with diverse texture distributions, volumetric intensity fields, implicit geometric surfaces, dynamic visual content, and spatial--angular LF data. The 1D experiments use the original KAN codebase, whereas the 2D--4D experiments use custom CUDA KAN layers integrated with PyTorch 2.0.0 and CUDA 11.3 on a single NVIDIA RTX 4090 GPU. All methods are optimized with AdamW. The 1D, 2D, and 4D tasks run for 5,000 iterations, while the 3D volume, SDF, and video tasks run for 20,000 iterations. Within each task, all methods use identical data, preprocessing, sampling, iteration budgets, and evaluation protocols, with method-specific settings following their official implementations.For the full model, $\lambda_{\mathrm{band}}=10^{-3}$ and $\alpha=0.5$; $\lambda_{\mathrm{band}}=0$ only in no-band ablations. Network parameters are optimized separately for each target signal, with audio fitted segment by segment as described in Section~\ref{sec:audio}.

\subsection{Parameterization of 1D Square-Wave Signals}

\subsubsection{Datasets}

Training uses 128 uniformly sampled square-wave points, while evaluation uses 256 points over the same domain to test reconstruction at additional coordinates not observed during optimization.

\subsubsection{Experimental Details}
All methods are optimized for 5,000 iterations with $\lambda_{\mathrm{band}}=10^{-3}$, $\alpha=0.5$, $J=3$, and the Haar wavelet.

\subsubsection{Experimental Results}

\begin{figure}[b]
\centering
\includegraphics[width=0.48\textwidth]{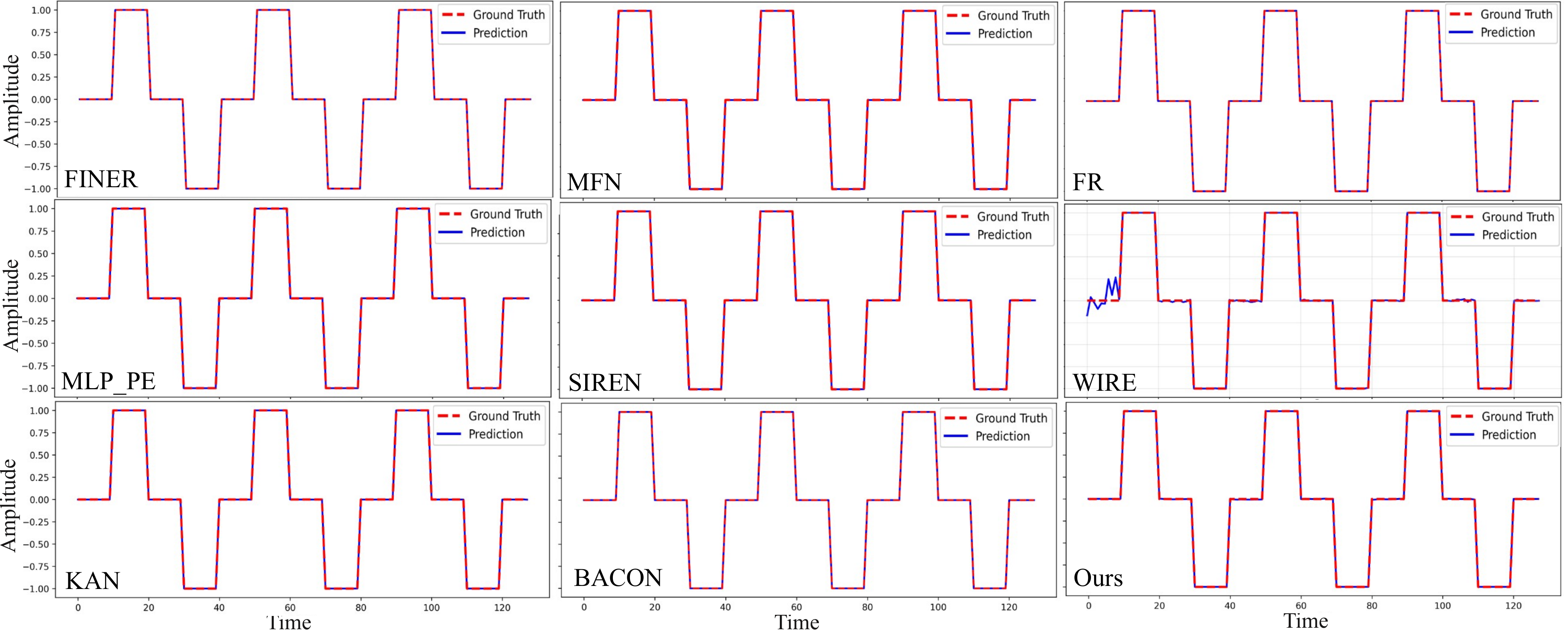}
\caption{Qualitative comparison of the 1D square-wave reconstruction results obtained by nine representation methods. The reference signal and the predictions of all methods are evaluated on the same coordinate grid.}
\label{fig-5}
\end{figure}

\begin{figure}[t]
\centering
\includegraphics[width=0.48\textwidth]{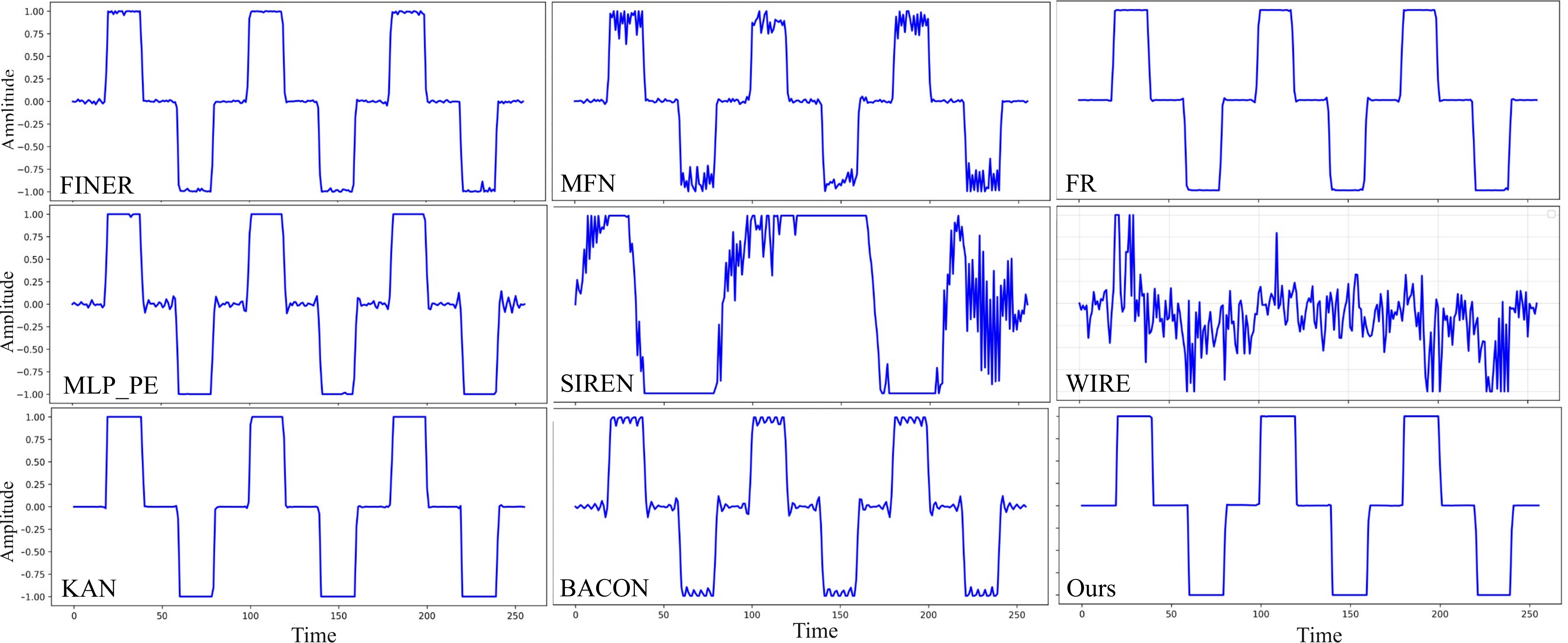}
\caption{Dense-grid evaluation of the square-wave representation results. All methods were optimized using 128 uniformly sampled coordinates and subsequently evaluated at 256 uniformly sampled coordinates over the same signal domain.}
\label{fig-6}
\end{figure}

Figs.~\ref{fig-5} and \ref{fig-6} present the square-wave reconstruction results on the original and denser coordinate grids. Square-wave signals contain piecewise-constant regions and abrupt transitions, providing a simple test of both smooth-region fitting and sharp-transition reconstruction. Several comparison methods exhibit transition blurring, oscillatory artifacts, or amplitude deviations near discontinuities, whereas the proposed method better preserves piecewise-constant regions and sharp transitions. On the denser coordinate grid, which contains additional coordinates not used during optimization, the proposed method also maintains more stable signal values and sharper transition profiles.

\subsection{Parameterization of 1D Audio Signals}
\label{sec:audio}

\subsubsection{Datasets}

For the 1D audio-fitting experiments, we used the same target audio clips as those used in SIREN~\cite{sitzmann2020implicit}, namely “bach” and “counting.” Their durations are 6.99 s and 12 s, respectively. Both clips are stored in WAV format at a sampling rate of 44.1 kHz. Each audio waveform was represented as a mapping from a normalized temporal coordinate to the corresponding signal amplitude.

\subsubsection{Experimental Details}

The audio experiments followed the optimization settings used in the 1D square-wave experiments. To reduce the difficulty of fitting long audio sequences and enable a consistent comparison among different representation methods, each audio waveform was divided into nonoverlapping 0.5-s segments. Each segment was independently parameterized using its corresponding temporal coordinates and amplitude values. After optimization, the reconstructed segments were concatenated in chronological order to obtain the complete audio waveform. All comparison methods used the same segmentation strategy and experimental settings.

\subsubsection{Experimental Results}

\begin{figure*}[!htp]
\centering
\includegraphics[width=0.95\textwidth]{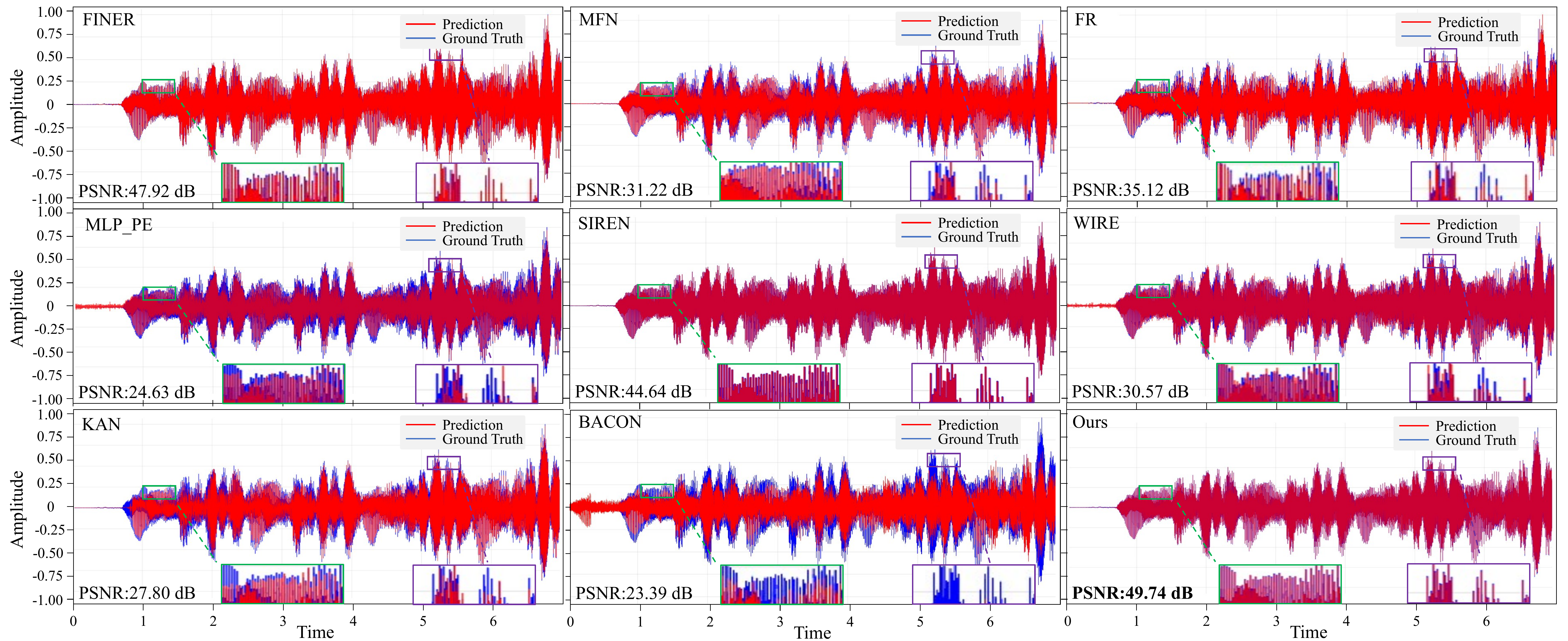}
\caption{Comparison of the 1D audio fitting results obtained by nine representation methods on the ``bach'' audio clip.}
\label{fig-3}
\end{figure*}

\begin{figure*}[!t]
\centering
\includegraphics[width=0.95\textwidth]{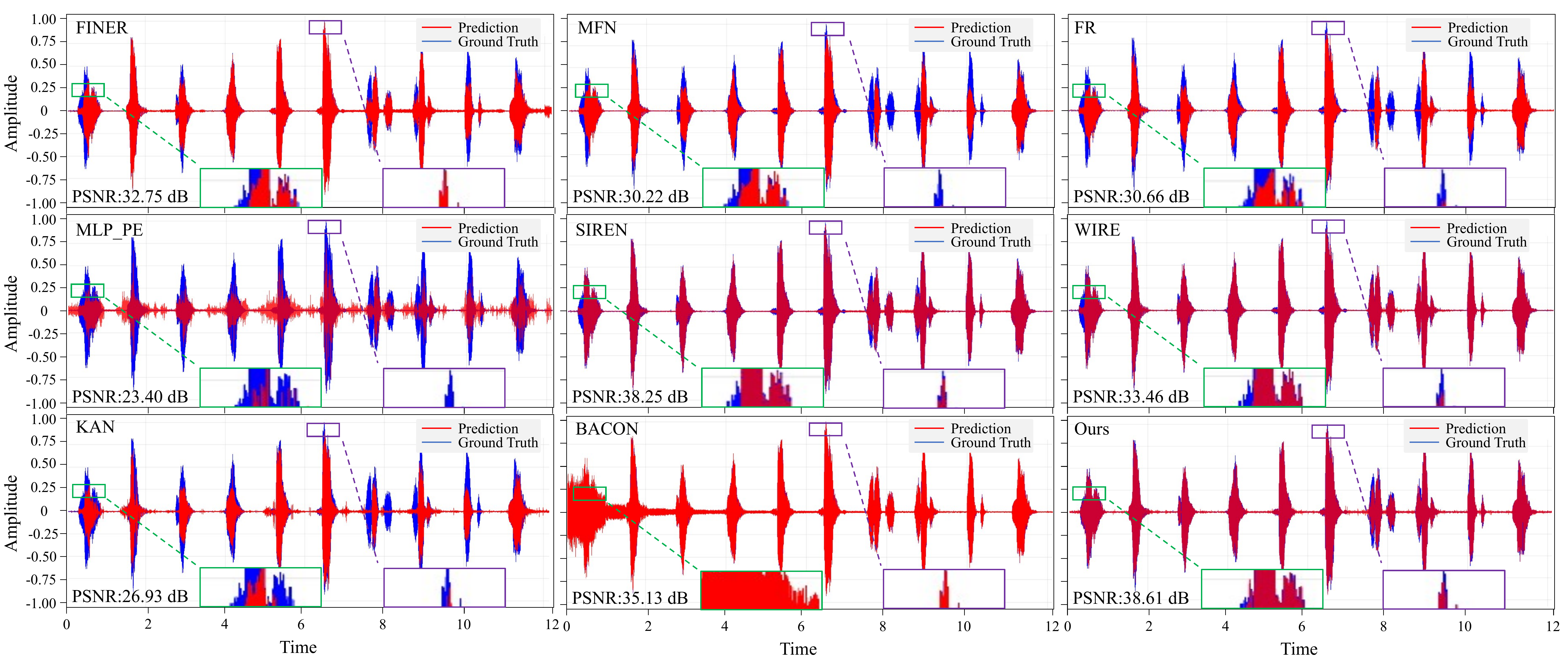}
\caption{Comparison of the 1D audio fitting results obtained by nine representation methods on the ``counting'' audio clip.}
\label{fig-4}
\end{figure*}

Figs.~\ref{fig-3} and \ref{fig-4} compare the audio fitting performance of the proposed method with that of eight representative coordinate-based methods. PSNR was used to quantify the sample-wise discrepancy between the reconstructed and reference audio waveforms. The two audio clips contain both slowly varying waveform structures and rapid local oscillations, thereby requiring the representation model to characterize signal components over different frequency ranges.

As shown in Figs.~\ref{fig-3} and \ref{fig-4}, several comparison methods produce noticeable reconstruction deviations in waveform regions containing rapid amplitude variations. In contrast, the proposed method more closely follows the reference waveform in both slowly varying and rapidly oscillating regions. The proposed method also achieves the highest PSNR among the compared methods on both audio clips, indicating improved sample-level reconstruction fidelity for 1D audio signals.

\subsection{Parameterization of 2D Images}
\label{sec:image_fitting}

\subsubsection{Datasets}

We use the first 700 images from the DIV2K Train Data (HR images) subset~\cite{agustsson2017ntire} for the 2D image parameterization experiments. All selected images are uniformly resized to $512\times512$ before training and evaluation. The selected images contain diverse scenes and exhibit different levels of structural and textural complexity. DIV2K is widely used for evaluating image reconstruction and super-resolution methods.

\subsubsection{Experimental Details}
All images are resized to $512\times512$ and evaluated using PSNR, SSIM, and LPIPS. The MLP uses three hidden layers of width 64, while the KAN widths are 128-256-128 with $k=3$, grid size 64, and $J=3$. This accuracy--efficiency configuration is used in Table~\ref{tab:wavelet_parameter_ablation} and subsequent multidimensional
experiments. Additional network settings are evaluated in the hyperparameter sensitivity analysis in Section~\ref{sec:ablation_hyperparameters}.

\subsubsection{Experimental Results}

\begin{figure*}[!t]
\centering
\includegraphics[width=0.95\textwidth]{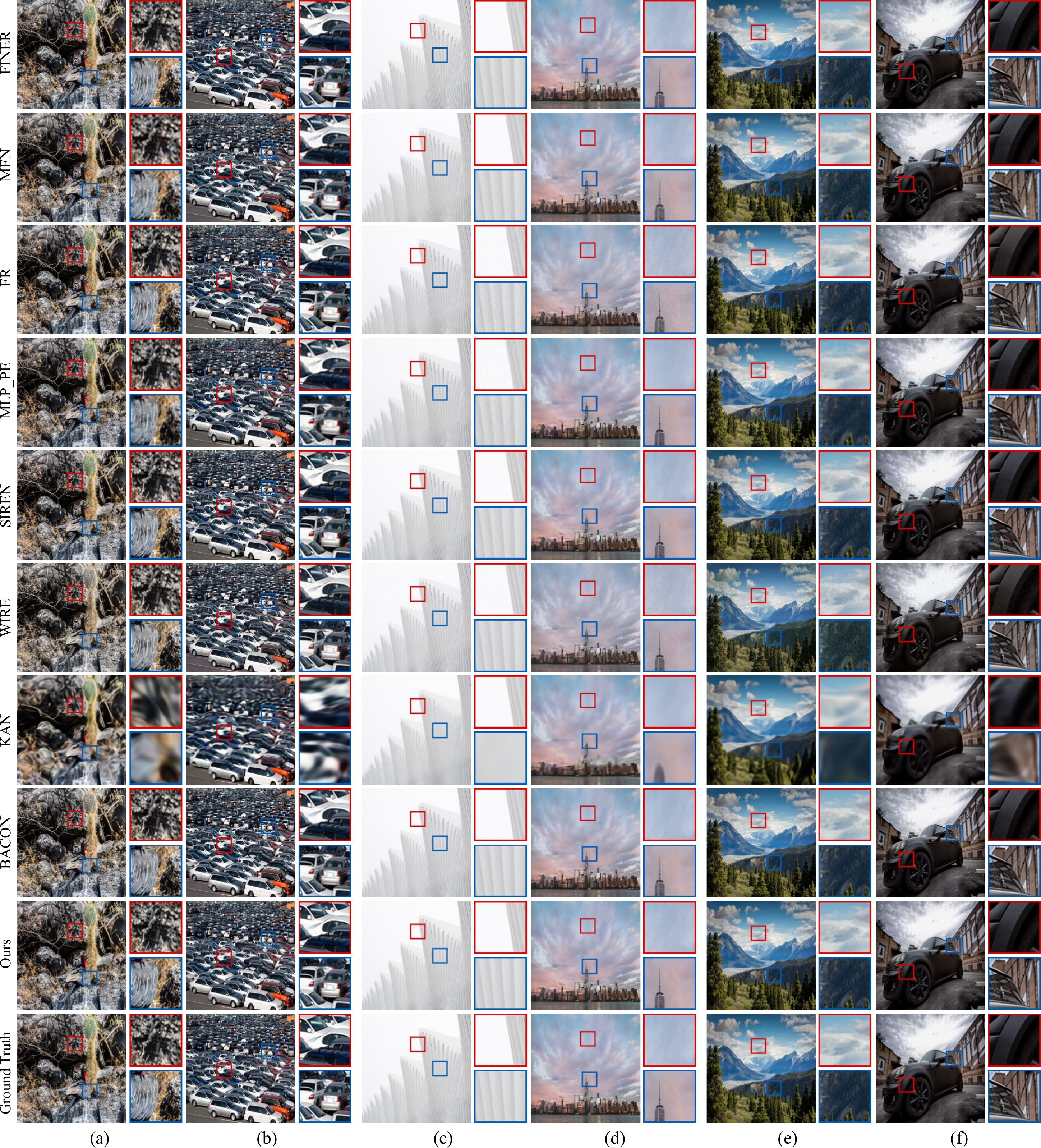}
\caption{Qualitative comparison of nine image representation methods on six representative DIV2K images with different levels of texture complexity. Subfigures (a) and (b) correspond to images with complex and dense textures, (c) and (d) correspond to images with relatively sparse textures, and (e) and (f) correspond to images containing both textured and smooth regions.}
\label{fig-7}
\end{figure*}

\begin{table}[!htp]
\centering
\caption{Image fitting performance on the DIV2K subset at a resolution of $512\times512$. Results are reported as mean $\pm$ standard deviation.}
\label{tab:div2k_average_results}
\scriptsize
\setlength{\tabcolsep}{7pt}
\renewcommand{\arraystretch}{1.5}
\begin{tabular}{@{}lccc@{}}
\toprule
Method & PSNR (dB) $\uparrow$ & SSIM $\uparrow$ & LPIPS $\downarrow$ \\
\midrule
FINER  & $39.36 \pm 3.837$  & $0.9726 \pm 0.0151$ & $0.0614 \pm 0.0338$ \\
MFN    & $31.24 \pm 4.334$  & $0.8887 \pm 0.0110$ & $0.2274 \pm 0.0121$ \\
FR     & $34.72 \pm 3.010$  & $0.9316 \pm 0.0389$ & $0.1308 \pm 0.0625$ \\
MLP-PE & $29.96 \pm 3.458$  & $0.8819 \pm 0.0434$ & $0.2374 \pm 0.0444$ \\
SIREN  & $37.48 \pm 3.291$  & $0.9781 \pm 0.0099$ & $0.0527 \pm 0.0253$ \\
WIRE   & $33.14 \pm 2.858$  & $0.8741 \pm 0.0481$ &
$0.2289 \pm 0.0686$ \\
KAN    & $26.89 \pm 2.782$  & $0.8106 \pm 0.0732$ & $0.3210 \pm 0.0694$ \\
BACON  & $30.81 \pm 2.571$  & $0.8920 \pm 0.0319$ & $0.2042 \pm 0.0590$ \\
\textbf{Ours}
& $\mathbf{45.72 \pm 2.349}$
& $\mathbf{0.9912 \pm 0.0030}$
& $\mathbf{0.0030 \pm 0.0020}$ \\
\bottomrule
\end{tabular}
\end{table}

Fig.~\ref{fig-7} shows that the proposed method better preserves fine textures and structural boundaries across images with different frequency characteristics. Table~\ref{tab:div2k_average_results} further shows the best average PSNR, SSIM, and LPIPS over 700 images; its PSNR of 45.72 dB exceeds the second-best FINER by 6.36 dB.

\subsection{Parameterization of 3D Volumes, SDFs, and Videos}

\begin{figure*}[!t]
\centering
\includegraphics[width=0.95\textwidth]{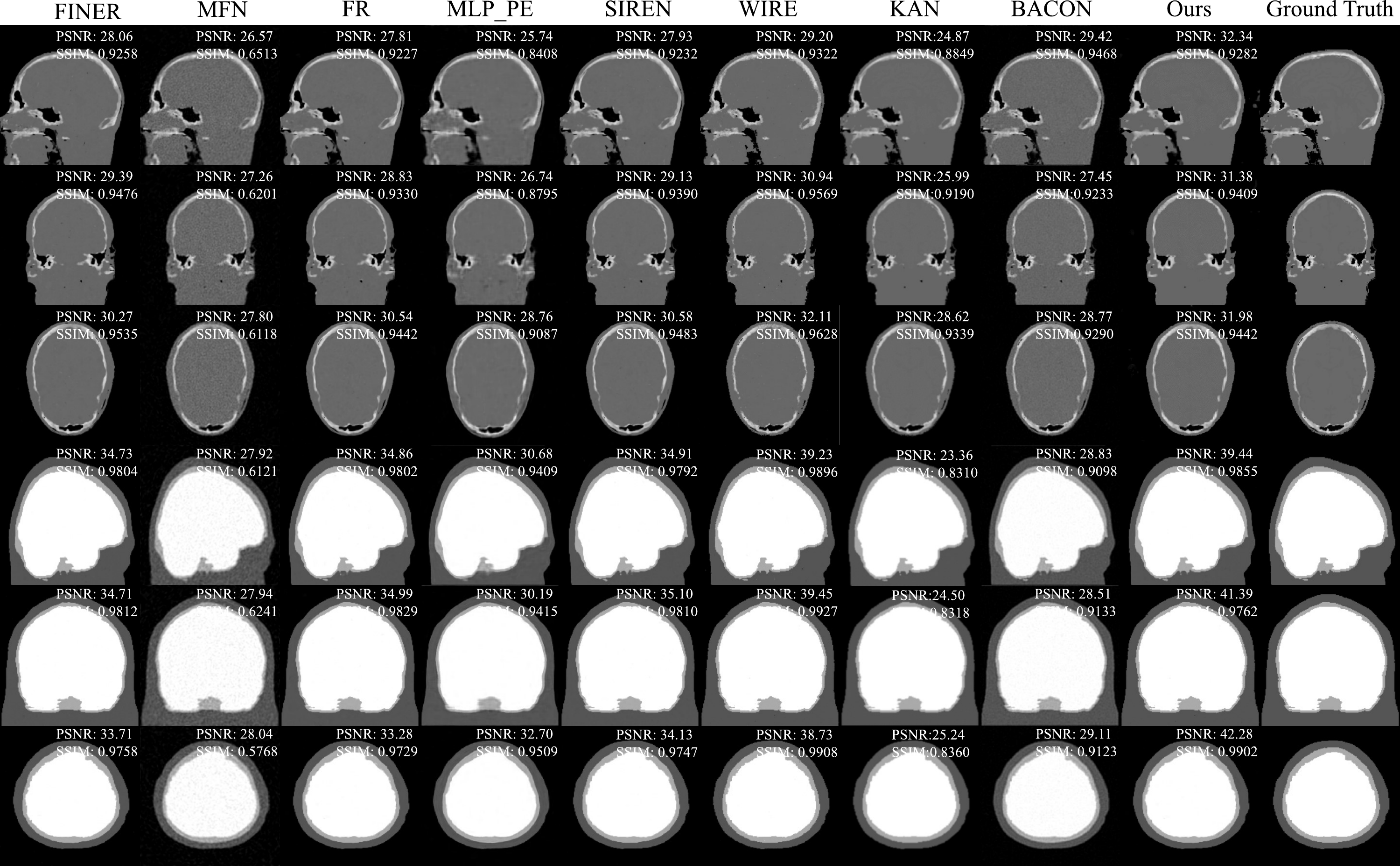}
\caption{Qualitative and quantitative comparison of volumetric reconstruction results on the Head and Standard datasets. Orthogonal cross-sections along the three spatial dimensions are shown, with the corresponding PSNR and SSIM values annotated for each method.}
\label{voxel}
\end{figure*}

\begin{figure}[!t]
\centering
\includegraphics[width=0.5\textwidth]{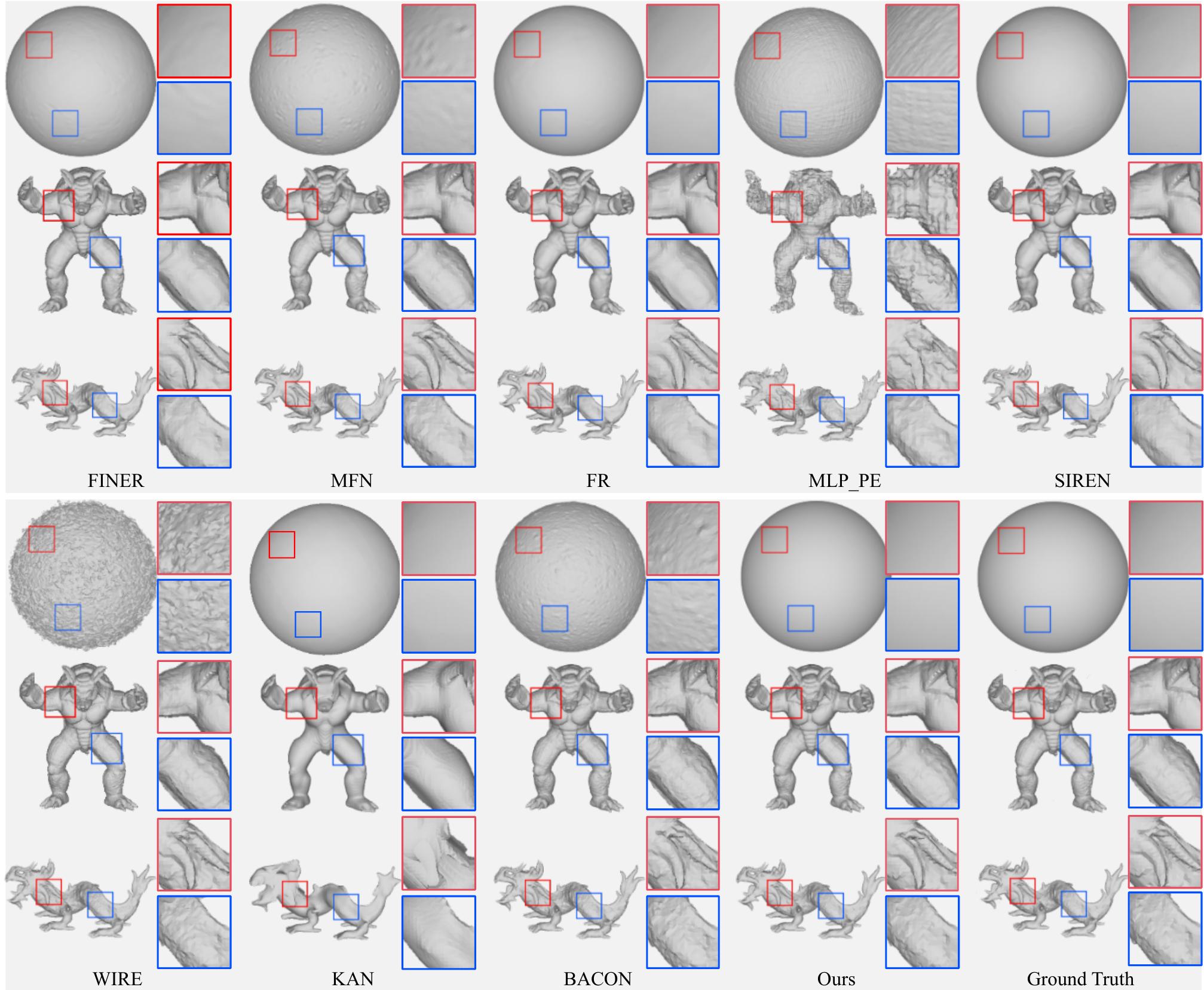}
\caption{Qualitative comparison of the reconstructed surfaces obtained by different representation methods on Sphere, Armadillo, and Dragon.}
\label{figsdf}
\end{figure}

\subsubsection{Datasets}

For volumetric fitting, we use the Head~\cite{biguri2016tigre} and Standard volumes~\cite{holmes1998enhancement}. SDF experiments use a synthetic Sphere and the Armadillo and Dragon models from the Stanford 3D Scanning Repository%
\footnote{\url{https://graphics.stanford.edu/data/3Dscanrep/}}, covering smooth and geometrically complex surfaces. Video experiments use the BALL, COLORS, and PARALLAX sequences from the MSU Video Compression Super-Resolution Benchmark~\cite{Bogatyrev_2024_BMVC}, with each sequence represented as a mapping from $(x,y,t)$ to RGB.

\subsubsection{Experimental Details}

The Head and Standard volumes have resolutions of $256\times256\times256$ and $181\times217\times181$, respectively. Video frames are resized to $480\times270$, with the first 100 frames used for fitting and evaluation. Volumetric, SDF, and video experiments follow the configuration in Section~\ref{sec:image_fitting} and are optimized for 20,000 iterations. Volumes and videos are evaluated using the corresponding PSNR/SSIM/LPIPS metrics, while SDF reconstruction is evaluated using Chamfer distance and IoU.

\subsubsection{Experimental Results}
\begin{table*}[!t]
\centering
\caption{Quantitative comparison on the Sphere, Armadillo, and Dragon SDF datasets. Results are reported as mean $\pm$ standard deviation over three runs. The best result for each metric is shown in bold. }
\label{tab:3d_sdf_results_mean_std}
\footnotesize
\setlength{\tabcolsep}{3pt}
\renewcommand{\arraystretch}{1.5}
\resizebox{\textwidth}{!}{%
\begin{tabular}{lcccccc}
\toprule
\multirow{2}{*}{Method}
& \multicolumn{2}{c}{Sphere}
& \multicolumn{2}{c}{Armadillo}
& \multicolumn{2}{c}{Dragon} \\
\cmidrule(lr){2-3}
\cmidrule(lr){4-5}
\cmidrule(lr){6-7}
& Chamfer $\downarrow$ & IoU $\uparrow$
& Chamfer $\downarrow$ & IoU $\uparrow$
& Chamfer $\downarrow$ & IoU $\uparrow$ \\
\midrule
    FINER
    & \meanstd{1.721\times10^{-4}}{2.52\times10^{-7}}
    & \meanstd{9.964\times10^{-1}}{3.61\times10^{-4}}
    & \meanstd{9.630\times10^{-6}}{4.86\times10^{-7}}
    & \meanstd{9.789\times10^{-1}}{1.19\times10^{-3}}
    & \meanstd{7.941\times10^{-6}}{5.66\times10^{-7}}
    & \meanstd{9.583\times10^{-1}}{7.02\times10^{-4}} \\
    MFN
    & \meanstd{1.735\times10^{-4}}{7.37\times10^{-7}}
    & \meanstd{9.967\times10^{-1}}{3.06\times10^{-4}}
    & \meanstd{8.192\times10^{-6}}{3.00\times10^{-8}}
    & \meanstd{9.924\times10^{-1}}{3.79\times10^{-4}}
    & \meanstd{6.556\times10^{-6}}{2.76\times10^{-8}}
    & \meanstd{9.758\times10^{-1}}{6.00\times10^{-4}} \\
    FR
    & \meanstd{1.730\times10^{-4}}{3.51\times10^{-7}}
    & \meanstd{9.970\times10^{-1}}{3.51\times10^{-4}}
    & \meanstd{9.333\times10^{-6}}{1.35\times10^{-7}}
    & \meanstd{9.814\times10^{-1}}{3.06\times10^{-4}}
    & \meanstd{8.076\times10^{-6}}{3.55\times10^{-7}}
    & \meanstd{9.570\times10^{-1}}{3.55\times10^{-3}} \\
    MLP-PE
    & \meanstd{1.739\times10^{-4}}{2.52\times10^{-7}}
    & \meanstd{9.931\times10^{-1}}{6.00\times10^{-4}}
    & \meanstd{4.697\times10^{-5}}{6.46\times10^{-5}}
    & \meanstd{9.514\times10^{-1}}{5.20\times10^{-2}}
    & \meanstd{7.406\times10^{-6}}{2.50\times10^{-7}}
    & \meanstd{9.628\times10^{-1}}{2.03\times10^{-3}} \\
    SIREN
    & \meanstd{1.722\times10^{-4}}{5.69\times10^{-7}}
    & \meanstd{9.959\times10^{-1}}{7.57\times10^{-4}}
    & \meanstd{1.032\times10^{-5}}{4.82\times10^{-7}}
    & \meanstd{9.761\times10^{-1}}{4.93\times10^{-4}}
    & \meanstd{9.438\times10^{-6}}{1.08\times10^{-6}}
    & \meanstd{9.480\times10^{-1}}{8.05\times10^{-3}} \\
    WIRE
    & \meanstd{2.230\times10^{-4}}{5.57\times10^{-7}}
    & \meanstd{9.742\times10^{-1}}{4.58\times10^{-4}}
    & \meanstd{9.555\times10^{-6}}{1.10\times10^{-8}}
    & \meanstd{9.920\times10^{-1}}{6.03\times10^{-4}}
    & \meanstd{1.515\times10^{-5}}{9.29\times10^{-8}}
    & \meanstd{9.624\times10^{-1}}{1.54\times10^{-3}} \\
    KAN
    & \meanstd{1.726\times10^{-4}}{6.08\times10^{-7}}
    & \meanstd{9.975\times10^{-1}}{4.00\times10^{-4}}
    & \meanstd{1.670\times10^{-5}}{6.71\times10^{-7}}
    & \meanstd{9.617\times10^{-1}}{6.03\times10^{-4}}
    & \meanstd{3.328\times10^{-5}}{5.87\times10^{-6}}
    & \meanstd{8.737\times10^{-1}}{1.25\times10^{-2}} \\
    BACON
    & \meanstd{1.741\times10^{-4}}{3.51\times10^{-7}}
    & \meanstd{9.953\times10^{-1}}{4.73\times10^{-4}}
    & \meanstd{8.205\times10^{-6}}{1.71\times10^{-8}}
    & \meanstd{9.913\times10^{-1}}{3.21\times10^{-4}}
    & \meanstd{6.564\times10^{-6}}{6.02\times10^{-8}}
    & \meanstd{9.749\times10^{-1}}{5.13\times10^{-4}} \\
    \textbf{Ours}
    & \meanstdbold{1.712\times10^{-4}}{6.51\times10^{-7}}
    & \meanstdbold{9.987\times10^{-1}}{3.21\times10^{-4}}
    & \meanstdbold{8.064\times10^{-6}}{1.06\times10^{-8}}
    & \meanstdbold{9.943\times10^{-1}}{2.07\times10^{-3}}
    & \meanstdbold{6.309\times10^{-6}}{4.71\times10^{-8}}
    & \meanstdbold{9.782\times10^{-1}}{1.83\times10^{-3}} \\
    \bottomrule
\end{tabular}}
\end{table*}

\begin{figure}[!t]
\centering
\includegraphics[width=0.5\textwidth]{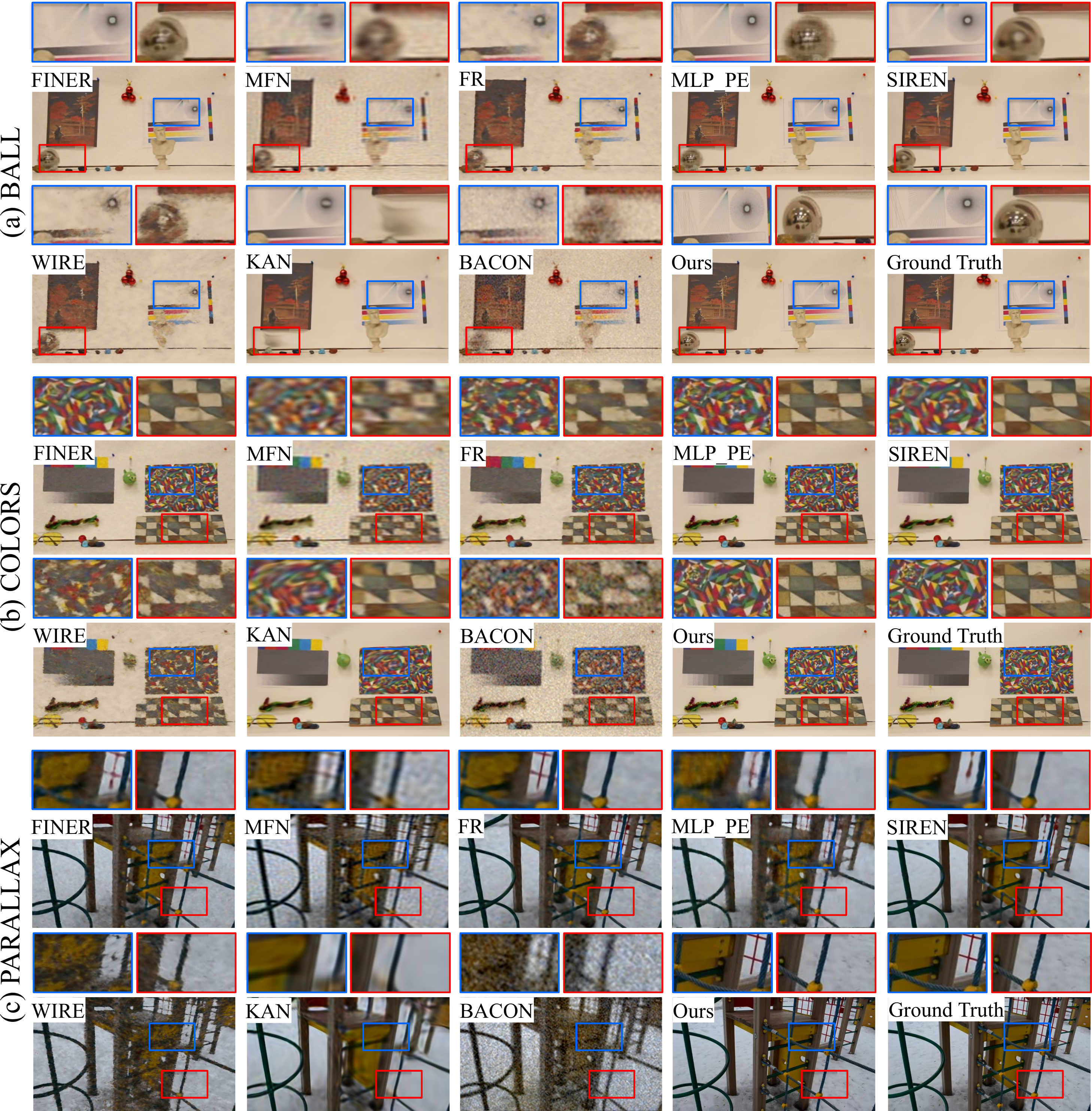}
\caption{Qualitative comparison of video reconstruction results on the BALL, COLORS, and PARALLAX sequences.}
\label{fig:video}
\end{figure}

\begin{table*}[!t]
\centering
\caption{Quantitative comparison on the BALL, COLORS, and PARALLAX video sequences. Results are reported as mean $\pm$ standard deviation over the evaluated frames. The best result for each metric is shown in bold.}
\label{tab:multi_scene_metric}
\small
\setlength{\tabcolsep}{5pt}
\renewcommand{\arraystretch}{1.5}
\resizebox{\textwidth}{!}{%
\begin{tabular}{lccccccccc}
\toprule
\multirow{2}{*}{Method}
& \multicolumn{3}{c}{BALL}
& \multicolumn{3}{c}{COLORS}
& \multicolumn{3}{c}{PARALLAX} \\
\cmidrule(lr){2-4}
\cmidrule(lr){5-7}
\cmidrule(lr){8-10}
& PSNR (dB) $\uparrow$ & SSIM $\uparrow$ & LPIPS $\downarrow$
& PSNR (dB) $\uparrow$ & SSIM $\uparrow$ & LPIPS $\downarrow$
& PSNR (dB) $\uparrow$ & SSIM $\uparrow$ & LPIPS $\downarrow$ \\
\midrule
    BACON
    & \(27.57 \pm 0.91\)
    & \(0.6811 \pm 0.0181\)
    & \(0.4827 \pm 0.0162\)
    & \(22.71 \pm 0.59\)
    & \(0.5066 \pm 0.0083\)
    & \(0.5888 \pm 0.0049\)
    & \(19.47 \pm 0.08\)
    & \(0.3445 \pm 0.0390\)
    & \(0.6601 \pm 0.0140\) \\
    FINER
    & \(36.99 \pm 0.49\)
    & \(0.9508 \pm 0.0315\)
    & \(0.1847 \pm 0.0108\)
    & \(30.34 \pm 0.14\)
    & \(0.9139 \pm 0.0174\)
    & \(0.2425 \pm 0.0226\)
    & \(27.01 \pm 0.27\)
    & \(0.7853 \pm 0.0409\)
    & \(0.3536 \pm 0.0169\) \\
    FR
    & \(29.52 \pm 0.23\)
    & \(0.8444 \pm 0.0130\)
    & \(0.4320 \pm 0.0218\)
    & \(25.93 \pm 0.62\)
    & \(0.8191 \pm 0.0030\)
    & \(0.3799 \pm 0.0290\)
    & \(26.53 \pm 0.28\)
    & \(0.8072 \pm 0.0336\)
    & \(0.3124 \pm 0.0188\) \\
    KAN
    & \(32.89 \pm 0.34\)
    & \(0.9390 \pm 0.0203\)
    & \(0.1670 \pm 0.0220\)
    & \(27.08 \pm 0.50\)
    & \(0.9025 \pm 0.0085\)
    & \(0.2158 \pm 0.0113\)
    & \(22.31 \pm 0.16\)
    & \(0.7167 \pm 0.0314\)
    & \(0.4413 \pm 0.0104\) \\
    MFN
    & \(28.08 \pm 0.95\)
    & \(0.7950 \pm 0.0082\)
    & \(0.4168 \pm 0.0100\)
    & \(24.60 \pm 0.19\)
    & \(0.7291 \pm 0.0238\)
    & \(0.4671 \pm 0.0146\)
    & \(23.26 \pm 0.36\)
    & \(0.6475 \pm 0.0102\)
    & \(0.4697 \pm 0.0061\) \\
    MLP-PE
    & \(38.89 \pm 0.86\)
    & \(0.9781 \pm 0.0282\)
    & \(0.1057 \pm 0.0050\)
    & \(31.11 \pm 0.24\)
    & \(0.9463 \pm 0.0448\)
    & \(0.1657 \pm 0.0196\)
    & \(23.63 \pm 0.26\)
    & \(0.7078 \pm 0.0122\)
    & \(0.4567 \pm 0.0174\) \\
    SIREN
    & \(35.04 \pm 0.35\)
    & \(0.9546 \pm 0.0111\)
    & \(0.1333 \pm 0.0280\)
    & \(29.13 \pm 0.25\)
    & \(0.9374 \pm 0.0216\)
    & \(0.1375 \pm 0.0063\)
    & \(26.14 \pm 0.47\)
    & \(\mathbf{0.8545 \pm 0.0308}\)
    & \(0.2665 \pm 0.0130\) \\
    WIRE
    & \(29.06 \pm 0.40\)
    & \(0.8318 \pm 0.0447\)
    & \(0.4126 \pm 0.0225\)
    & \(24.26 \pm 0.22\)
    & \(0.7598 \pm 0.0460\)
    & \(0.4539 \pm 0.0167\)
    & \(23.36 \pm 0.10\)
    & \(0.6740 \pm 0.0375\)
    & \(0.4710 \pm 0.0232\) \\
    \textbf{Ours}
    & \(\mathbf{43.77 \pm 0.53}\)
    & \(\mathbf{0.9883 \pm 0.0018}\)
    & \(\mathbf{0.0069 \pm 0.0009}\)
    & \(\mathbf{32.77 \pm 0.87}\)
    & \(\mathbf{0.9523 \pm 0.0146}\)
    & \(\mathbf{0.0369 \pm 0.0044}\)
    & \(\mathbf{27.42 \pm 0.10}\)
    & \(0.7744 \pm 0.0107\)
    & \(\mathbf{0.2211 \pm 0.0091}\) \\
    \bottomrule
\end{tabular}}
\end{table*}

\paragraph{Volumetric representation}
Fig.~\ref{voxel} compares the reconstructed cross-sections of the Head and Standard volumes. The competing methods exhibit different degrees of smoothing and structural deviation near tissue boundaries and small anatomical regions. The proposed method preserves the principal anatomical structures while producing clearer boundaries and fewer local reconstruction artifacts. These results demonstrate that the proposed representation can model both slowly varying volumetric regions and localized structural transitions.

\paragraph{SDF representation}
Fig.~\ref{figsdf} and Table~\ref{tab:3d_sdf_results_mean_std} show that the proposed method achieves the lowest mean Chamfer distance and highest mean IoU on all three shapes, indicating consistent reconstruction of both smooth and geometrically complex surfaces.

\paragraph{Video representation}
Fig.~\ref{fig:video} and Table~\ref{tab:multi_scene_metric} show that the proposed method achieves the best result in eight of the nine reported metrics, including the highest PSNR and lowest LPIPS on all three sequences. The only exception is SSIM on PARALLAX, where SIREN achieves 0.8545 compared with 0.7744 for the proposed method.

\subsection{Parameterization of 4D Light Fields}

\subsubsection{Datasets}

Four scenes from the HCI 4D LF Benchmark~\cite{honauer2016dataset} are used. Each scene contains $9\times9$ regularly arranged subaperture images, and all 81 views are used for fitting and evaluation.

\subsubsection{Experimental Details}

Each LF is represented as $(x,y,u,v)\mapsto RGB$. During optimization, spatial patches and their corresponding 4D coordinates are randomly sampled across angular views. All methods use identical views, sampling, and optimization conditions, follow the settings in Section~\ref{sec:experimental_setup}, and are trained for 5,000 iterations. PSNR, SSIM, and LPIPS are averaged over all reconstructed subaperture views.

\subsubsection{Experimental Results}

\begin{table}[t]
\centering
\caption{Quantitative comparison of different implicit neural representation methods on four HCI LF scenes. Results are reported as mean $\pm$ standard deviation. The best result for each metric is shown in bold.}
\label{tab:lf_metric}
\footnotesize
\setlength{\tabcolsep}{4pt}
\renewcommand{\arraystretch}{1.5}
\begin{tabular}{@{}lccc@{}}
\toprule
Method & PSNR (dB) $\uparrow$ & SSIM $\uparrow$ & LPIPS $\downarrow$ \\
\midrule
BACON   & $25.51 \pm 1.130$ & $0.6568 \pm 0.0694$ & $0.5721 \pm 0.0821$ \\
FINER   & $24.42 \pm 1.872$ & $0.5796 \pm 0.0956$ & $0.4672 \pm 0.0809$ \\
FR      & $27.57 \pm 1.160$ & $0.7246 \pm 0.0753$ & $0.4382 \pm 0.1056$ \\
KAN     & $27.98 \pm 1.259$ & $0.7382 \pm 0.0673$ & $0.3644 \pm 0.1034$ \\
MFN     & $25.68 \pm 1.065$ & $0.6559 \pm 0.0741$ & $0.5344 \pm 0.0976$ \\
MLP-PE  & $28.89 \pm 1.590$ & $0.7568 \pm 0.0737$ & $\mathbf{0.2803 \pm 0.0940}$ \\
SIREN   & $28.60 \pm 1.218$ & $0.7576 \pm 0.0714$ & $0.3805 \pm 0.0107$ \\
WIRE    & $23.43 \pm 1.271$ & $0.5353 \pm 0.0446$ & $0.6112 \pm 0.0224$ \\
\textbf{Ours}
& $\mathbf{29.06 \pm 2.533}$
& $\mathbf{0.8102 \pm 0.0697}$
& $ 0.3607 \pm 0.0790$ \\
\bottomrule
\end{tabular}
\end{table}

\begin{figure*}[!hpt]
\centering
\includegraphics[width=0.95\textwidth]{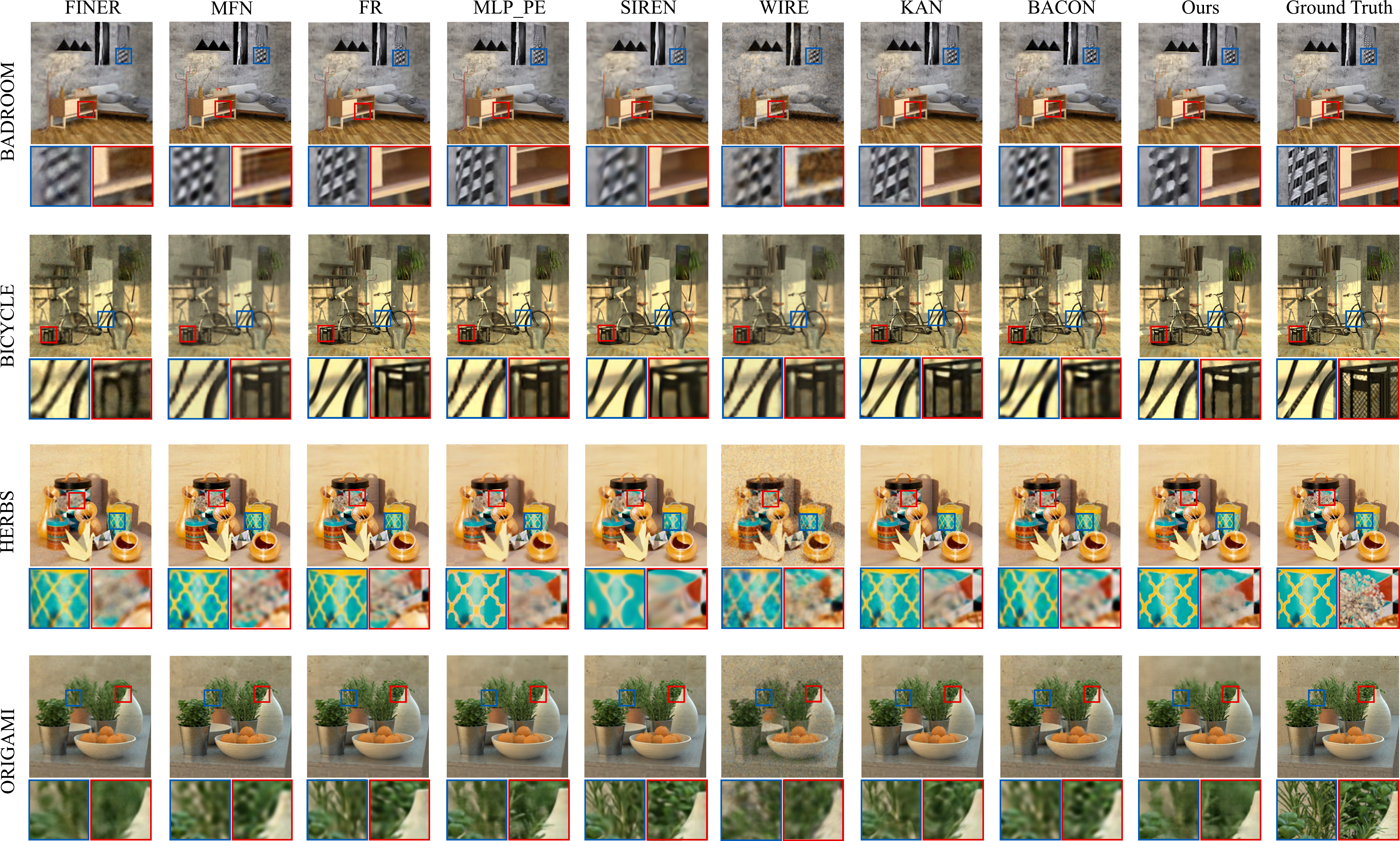}
\caption{Qualitative comparison of the 4D LF reconstruction results obtained by different INR-based methods on four representative HCI scenes. Representative subaperture views and enlarged local regions are shown.}
\label{fig:light_field_vis}
\end{figure*}

Fig.~\ref{fig:light_field_vis} shows that the proposed method better preserves scene structures and local textures in the representative subaperture views. Table~\ref{tab:lf_metric} shows that the proposed method achieves the highest average PSNR and SSIM over the four scenes, while MLP-PE achieves the lowest LPIPS. The proposed method improves the average PSNR by 0.17~dB over MLP-PE, which achieves the second-highest PSNR.

\begin{table}[!b]
    \centering
    \caption{Computational efficiency comparison across 1D--4D signal-fitting tasks using a batch size of 2048.}
    \label{tab:computational_efficiency_all_tasks}
    \scriptsize
    \setlength{\tabcolsep}{5.5pt}
    \renewcommand{\arraystretch}{0.95}
    \begin{tabular}{@{}clcccc@{}}
        \toprule
        \textbf{Task}& \textbf{Method}& \textbf{Params.}& \textbf{FLOPs}& \textbf{Time (s)}& \textbf{Mem. (MiB)} \\
        \midrule
        \multirow{9}{*}{\rotatebox[origin=c]{90}{\textbf{1D square wave}}}
        & FINER   & 198K  & 0.85G  & 0.4   & 46.5   \\
        & MFN     & 201K  & 0.91G  & 2.0    & 74.8   \\
        & FR      & 1.57M & 1.65G & 0.5    & 61.3   \\
        & MLP-PE  & 332K  & 1.42G  & 0.8    & 48.4   \\
        & SIREN   & 329K  & 1.41G  & 0.3    & 48.3   \\
        & WIRE    & 264K  & 2.27G & 0.5    & 64.3  \\
        & KAN     & 869K  & 3.95G  & 9.2    & 1261.4   \\
        & BACON   & 267K  & 1.13G  & 1.4    & 33.0   \\
        & Ours    & 1.97M & 0.55G & 1.7    & 207.6   \\
        \midrule
        \multirow{9}{*}{\rotatebox[origin=c]{90}{\textbf{2D image}}}
        & FINER   & 198K  & 0.86G  & 0.9    & 46.1 \\
        & MFN     & 201K  & 0.26G   & 7.9    & 46.8  \\
        & FR      & 1.57M &1.66G  & 1.9    & 63.1  \\
        & MLP-PE  & 332K  & 0.88G  & 20.9    & 40.9  \\
        & SIREN   & 329K  & 1.13G  & 3.8    & 45.3  \\
        & WIRE    & 264K  & 2.86G  & 18.1   & 66.1 \\
        & KAN     & 136K  &0.69G  & 17.3  & 390.8  \\
        & BACON   & 405K  & 1.72G  & 1.6    & 80.7 \\
        & Ours    & 4.13M & 20.51G  & 27.7   & 108.6  \\
        \midrule
        \multirow{9}{*}{\rotatebox[origin=c]{90}{\textbf{3D voxel}}}
        & FINER   & 264K  & 1.08G & 25.7   & 101.3 \\
        & MFN     & 274K  & 1.12G & 334.1   & 101.3 \\
        & FR      & 2.10M & 2.15G & 26.6   & 112.3  \\
        & MLP-PE  & 180K  & 0.86G  & 186.9   & 97.1  \\
        & SIREN   & 264K  & 1.07G & 15.9   & 97.3  \\
        & WIRE    & 265K  & 0.53G  & 269.0  & 111.0 \\
        & KAN     & 449K  & 5.13G  & 349.2 & 608.5 \\
        & BACON   & 529K  & 1.09G &223.5   & 101.6 \\
        & Ours    & 4.56M &19.09G & 136.6 & 954.8  \\
        \midrule
        \multirow{9}{*}{\rotatebox[origin=c]{90}{\textbf{4D LF}}}
        & FINER   & 199K  & 0.86G & 1.1   & 56.64  \\
        & MFN     & 228K  & 0.94G  & 24.5   & 75.2  \\
        & FR      & 1.58M & 1.66G  & 1.4   & 61.4  \\
        & MLP-PE  & 363K  & 0.90G  & 1.0   & 33.8  \\
        & SIREN   & 256K  & 1.14G  & 1.7   & 43.6  \\
        & WIRE    & 266K  & 27.10G  & 47.5   & 76.4  \\
        & KAN     & 142K  & 3.65G  & 37.6 &355.0  \\
        & BACON   & 266K  & 1.73G  & 2.2   & 77.2  \\
        & Ours    & 4.42M & 11.43G  & 47.1   & 65.6  \\
        \bottomrule
    \end{tabular}
\end{table}

\subsection{Computational Efficiency Analysis}

Table~\ref{tab:computational_efficiency_all_tasks} compares parameter count, FLOPs, convergence time, and peak GPU memory using a batch size of 2048. Convergence time denotes the elapsed optimization time required to reach 20~dB for the 1D and 2D tasks, 24~dB for the 3D voxel task, and 18~dB for the 4D LF  task. Because each INR is optimized independently for a target signal, we emphasize optimization convergence time rather than inference time. The proposed framework generally uses more parameters than the single-branch baselines and introduces additional computational cost in the higher-dimensional settings, reflecting an accuracy--computation trade-off. The standalone KAN baseline uses the original PyTorch implementation, whereas the proposed 2D--4D model uses custom CUDA KAN layers. Therefore, the reported convergence times involving KAN should be interpreted as implementation-dependent results.

\section{Ablation Experiments}

\subsection{Ablation on Network Hyperparameters}
\label{sec:ablation_hyperparameters}

\begin{table}[t]
\centering
\caption{Hyperparameter sensitivity analysis on the first 100 DIV2K images at $512\times512$. Each block represents an independent hyperparameter sweep. The nominal settings are MLP $3\times256$, KAN $3\times128$, $k=3$, and grid size 64.}
\label{tab:ablation}
\scriptsize
\setlength{\tabcolsep}{4pt}
\renewcommand{\arraystretch}{1.25}
\begin{tabular}{@{}ccccc@{}}
\toprule
& \textbf{Setting} & \textbf{PSNR (dB)} $\uparrow$ & \textbf{SSIM} $\uparrow$ & \textbf{LPIPS} $\downarrow$ \\
\midrule

\multirow[c]{3}{*}{%
    \rotatebox[origin=c]{90}{\textbf{Grid}}%
}
& $32$ & $39.98 \pm 3.64$ & $0.9785 \pm 0.0090$ & $0.0076 \pm 0.0057$ \\

& $64$ & $\mathbf{50.77 \pm 3.46}$ & $\mathbf{0.9974 \pm 0.0010}$ & $\mathbf{0.0005 \pm 0.0003}$ \\

& $128$ & $42.88 \pm 5.06$ & $0.9814 \pm 0.0094$ & $0.0072 \pm 0.0079$ \\

\midrule

\multirow[c]{3}{*}{%
    \rotatebox[origin=c]{90}{\textbf{Degree}}%
}
& $k=1$ & $47.13 \pm 2.52$ & $0.9938 \pm 0.0042$ & $0.0015 \pm 0.0012$ \\

& $k=2$ & $49.92 \pm 2.28$ & $0.9970 \pm 0.0014$ & $0.0006 \pm 0.0004$ \\

& $k=3$ & $\mathbf{50.90 \pm 3.35}$ & $\mathbf{0.9975 \pm 0.0008}$ & $\mathbf{0.0005 \pm 0.0004}$ \\

\midrule

\multirow[c]{12}{*}{%
    \rotatebox[origin=c]{90}{\textbf{KAN} ($L\times W$)}%
}
& $1 \times 64$ & $22.89 \pm 2.73$ & $0.5579 \pm 0.0950$ & $0.5566 \pm 0.0590$ \\
& $2 \times 64$ & $28.86 \pm 3.11$ & $0.8081 \pm 0.0460$ & $0.2034 \pm 0.0530$ \\
& $3 \times 64$ & $35.29 \pm 3.88$ & $0.9394 \pm 0.0240$ & $0.0480 \pm 0.0250$ \\
& $4 \times 64$ & $39.24 \pm 3.75$ & $0.9701 \pm 0.0120$ & $0.0203 \pm 0.0480$ \\
& $1 \times 128$ & $30.21 \pm 3.57$ & $0.8540 \pm 0.0490$ & $0.1299 \pm 0.0380$ \\
& $2 \times 128$ & $31.74 \pm 2.71$ & $0.9345 \pm 0.0010$ & $0.0080 \pm 0.0131$ \\
& $3 \times 128$ & $45.59 \pm 4.84$ & $0.9916 \pm 0.0060$ & $0.0056 \pm 0.0070$ \\
& $4 \times 128$ & $47.08 \pm 5.55$ & $0.9927 \pm 0.0050$ & $0.0029 \pm 0.0070$ \\
& $1 \times 256$ & $24.08 \pm 1.11$ & $0.6875 \pm 0.1030$ & $0.4143 \pm 0.0780$ \\
& $2 \times 256$ & $49.66 \pm 6.02$ & $0.9939 \pm 0.0050$ & $0.0006 \pm 0.0030$ \\
& $3 \times 256$ & $\mathbf{66.22 \pm 5.28}$ & $\mathbf{0.9998 \pm 0.0003}$ & $\mathbf{0.00004 \pm 0.0001}$ \\

& $4 \times 256$ & $56.51 \pm 4.02$ & $0.9985 \pm 0.0010$ & $0.0006 \pm 0.0010$ \\

\midrule

\multirow[c]{12}{*}{%
    \rotatebox[origin=c]{90}{\textbf{MLP} ($L\times W$)}%
}
& $1 \times 64$& $42.62 \pm 1.97$& $0.9920 \pm 0.0013$& $0.0010 \pm 0.0003$ \\
& $2 \times 64$& $50.25 \pm 2.15$& $0.9972 \pm 0.0009$& $0.0005 \pm 0.0003$ \\
& $3 \times 64$& $50.89 \pm 3.39$& $0.9974 \pm 0.0008$& $0.0004 \pm 0.0003$ \\
& $4 \times 64$& $51.03 \pm 3.38$& $0.9975 \pm 0.0007$& $0.0004 \pm 0.0004$ \\
& $1 \times 128$& $44.34 \pm 1.69$ & $0.9923 \pm 0.0002$& $0.0018 \pm 0.0004$ \\
& $2 \times 128$& $50.87 \pm 3.16$& $0.9975 \pm 0.0006$& $0.0005 \pm 0.0005$ \\
& $3 \times 128$& $50.64 \pm 3.20$& $0.9973 \pm 0.0008$& $0.0005 \pm 0.0005$ \\
& $4 \times 128$& $51.21 \pm 3.56$& $0.9975 \pm 0.0008$& $0.0004 \pm 0.0004$ \\
& $1 \times 256$& $44.50 \pm 2.23$& $0.9925 \pm 0.0009$& $0.0009 \pm 0.0004$ \\
& $2 \times 256$& $50.89 \pm 3.41$& $0.9974 \pm 0.0007$& $0.0005 \pm 0.0003$ \\
& $3 \times 256$& $50.52 \pm 2.31$& $0.9974 \pm 0.0008$& $0.0004 \pm 0.0003$ \\
& $4 \times 256$& $\mathbf{51.29 \pm 4.09}$& $\mathbf{0.9975 \pm 0.0006}$& $\mathbf{0.0004 \pm 0.0002}$ \\
\bottomrule
\end{tabular}
\vspace{1mm}
\begin{minipage}{0.78\textwidth}
\footnotesize
\end{minipage}
\end{table}

We evaluate the sensitivity to KAN grid size, B-spline degree, KAN architecture, and MLP architecture on the first 100 DIV2K images at $512\times512$. Each block in Table~\ref{tab:ablation} is conducted as an independent hyperparameter sweep. Within each block, only the factor under study is varied while the remaining settings are fixed. The nominal settings are MLP $3\times256$, KAN $3\times128$, $k=3$, and grid size 64. These nominal settings are used only for the independent sensitivity sweeps and are distinct from the accuracy--efficiency configuration adopted in the main experiments. The KAN-architecture sweep uses our CUDA-fused implementation and is therefore mainly interpreted through within-block comparisons.

\begin{table}[t]
    \centering
    \caption{Sensitivity analysis of the wavelet level $J$ and Daubechies basis on the first 100 DIV2K images at $512\times512$. The adopted configuration uses an MLP with three hidden layers of width 64 and a KAN with widths 128-256-128, $k=3$, and grid size 64.}
    \label{tab:wavelet_parameter_ablation}
    \scriptsize
    \setlength{\tabcolsep}{2pt}
    \renewcommand{\arraystretch}{1.25}
    \resizebox{\columnwidth}{!}{%
    \begin{tabular}{@{}lccc@{}}
        \toprule
        \textbf{Setting}
        & \textbf{PSNR (dB)} $\uparrow$
        & \textbf{SSIM} $\uparrow$
        & \textbf{LPIPS} $\downarrow$ \\
        \midrule
        $J=1$, db1        & $45.1160 \pm 2.3574$        & $0.9910 \pm 0.0036$        & $0.0031 \pm 0.0023$ \\
        $J=1$, db2        & $45.1463 \pm 2.3610$        & $0.9911 \pm 0.0030$        & $0.0031 \pm 0.0020$ \\
        $J=1$, db3        & $45.1390 \pm 2.3730$        & $0.9911 \pm 0.0031$        & $0.0030 \pm 0.0022$ \\
        $J=1$, db4        & $45.1068 \pm 2.3415$        & $0.9910 \pm 0.0030$        & $0.0030 \pm 0.0021$ \\
        \midrule
        $J=2$, db1        & $45.1527 \pm 2.3704$        & $0.9911 \pm 0.0031$        & $0.0030 \pm 0.0016$ \\
        $J=2$, db2        & $45.1226 \pm 2.3318$        & $0.9910 \pm 0.0031$        & $0.0031 \pm 0.0016$ \\
        $J=2$, db3        & $45.1667 \pm 2.3600$        & $0.9911 \pm 0.0031$        & $0.0031 \pm 0.0017$ \\
        $J=2$, db4        & $45.1473 \pm 2.3033$        & $0.9910 \pm 0.0031$        & $0.0031 \pm 0.0017$ \\
        \midrule
        \rowcolor{gray!10}
        $J=3$, db1        & $45.1687 \pm 2.332$        & $0.9911 \pm 0.0030$        & $0.0032 \pm 0.0020$ \\
        $J=3$, db2        & $45.1698 \pm 2.370$        & $0.9910 \pm 0.0030$        & $0.0031 \pm 0.0020$ \\
        $J=3$, db3        & $45.1747 \pm 2.302$        & $0.9911 \pm 0.0030$        & $0.0032 \pm 0.0020$ \\
        $J=3$, db4        & $45.1700 \pm 2.324$        & $0.9910 \pm 0.0030$        & $0.0031 \pm 0.0020$ \\
        \midrule
        $J=4$, db1        & $45.1900 \pm 2.3378$        & $0.9910 \pm 0.0032$        & $0.0031 \pm 0.0017$ \\
        $J=4$, db2        & $45.1973 \pm 2.3515$        & $0.9911 \pm 0.0031$        & $0.0031 \pm 0.0017$ \\
        $J=4$, db3& $45.2157 \pm 2.3514$ & $0.9911 \pm 0.0032$& $0.0031 \pm 0.0017$ \\
        $J=4$, db4& $45.1640 \pm 2.3371$& $0.9910 \pm 0.0032$& $0.0031 \pm 0.0018$ \\

        \bottomrule
    \end{tabular}%
    }
\end{table}

The KAN branch is highly sensitive to its architecture, while increasing network depth or width does not necessarily improve reconstruction performance. In contrast, the MLP branch remains relatively stable across different depths and widths. We further evaluate the sensitivity to the wavelet decomposition level $J$ and basis on the first 100 DIV2K images at $512\times512$. Table~\ref{tab:wavelet_parameter_ablation} reports the results under the accuracy--efficiency configuration described in Section~\ref{sec:image_fitting}. The evaluated settings exhibit only minor performance variations. Therefore, we adopt $J=3$ and db1 as the wavelet configuration in the main experiments.

\subsection{Ablation on Band-Separation Regularization}

\begin{figure*}[!h]
\centering
\includegraphics[width=0.95\textwidth]{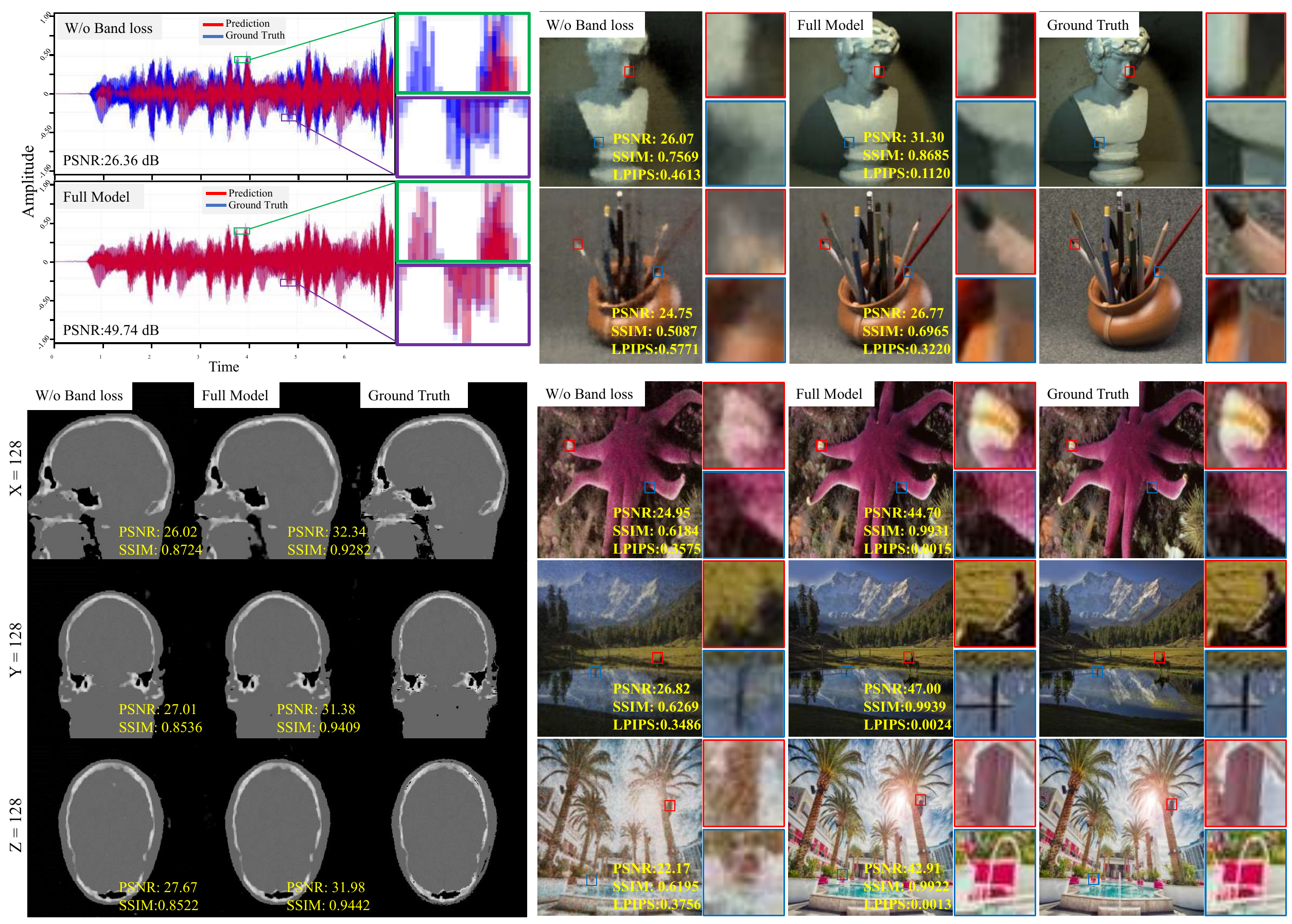}
\caption{Ablation results for the band-separation regularization on 1D audio, 2D images, 3D volumetric data, and 4D LFs. The full model is compared with a variant in which $\mathcal{L}_{\mathrm{band}}$ is removed while DWT/IDWT and additive wavelet-domain fusion are retained. Quantitative metrics are reported in the corresponding subfigures.}
\label{fig:all_dim_ablation}
\end{figure*}

To isolate the effect of the band-separation regularization, we compare the full model with a variant denoted W/o Band Loss on 1D audio, 2D image, 3D volume, and 4D LF fitting tasks. Both variants retain the same DWT/IDWT operations and additive wavelet-domain fusion, while $\mathcal{L}_{\mathrm{band}}$ is removed only in the ablated variant. All other network and optimization settings remain unchanged. As shown in Fig.~\ref{fig:all_dim_ablation}, removing $\mathcal{L}_{\mathrm{band}}$ consistently degrades reconstruction quality across the evaluated tasks, demonstrating its contribution to frequency-oriented coordination and reconstruction fidelity.

To further examine the interaction between the fusion rule and band-separation regularization, we conduct a controlled study on 100 DIV2K images using four variants: additive fusion with and without $\mathcal{L}_{\mathrm{band}}$, and fixed routing with and without $\mathcal{L}_{\mathrm{band}}$. All other settings are kept identical. Additive + Band corresponds to the full model, whereas Additive + No Band uses the same 2D setting as the W/o Band variant in Fig.~\ref{fig:all_dim_ablation}. The fixed-routing variants are introduced only as controlled baselines that explicitly impose the ideal frequency-selective assignment described in Eq.~\eqref{eq:frequency_assignment}, rather than as the forward mechanism of the proposed model.

\begin{table*}[t]
\centering
\caption{Controlled analysis of the fusion strategy and band separation regularization on 100 DIV2K images. MLP-L/H and KAN-L/H denote the low- and high-frequency energy proportions normalized within the corresponding branch. Results are reported as mean $\pm$
standard deviation.}
\label{tab:fusion_band_ablation}
\scriptsize
\setlength{\tabcolsep}{3pt}
\renewcommand{\arraystretch}{1.5}
\begin{tabular*}{\textwidth}
{@{\extracolsep{\fill}}lccc|cccc@{}}
\hline
\multirow{2}{*}{Variant}
& \multicolumn{3}{c|}{Reconstruction Performance}
& \multicolumn{4}{c}{Branch-wise Wavelet Energy (\%)} \\
\cline{2-8}
& PSNR (dB) $\uparrow$
& SSIM $\uparrow$
& LPIPS $\downarrow$
& MLP-L
& MLP-H
& KAN-L
& KAN-H \\
\hline

\textbf{Additive + Band}
& \textbf{44.60$\pm$2.65}
& \textbf{0.9908$\pm$0.0027}
& \textbf{0.0031$\pm$0.0017}
& (99.99993$\pm$3.0)$\times10^{-5}$
& $(7.5\pm3.0)\times10^{-5}$
& 84.34$\pm$6.80
& 15.67$\pm$6.80 \\

Additive + No Band
& 28.93$\pm$2.79
& 0.6914$\pm$0.0027
& 0.0397$\pm$0.0015
& (99.99981$\pm$8.0)$\times10^{-5}$
& (1.86$\pm$0.80)$\times10^{-4}$
& 96.00$\pm$2.90
& 4.00$\pm$2.90 \\

Fixed + Band
& 19.21$\pm$1.64
& 0.8628$\pm$0.0451
& 0.1758$\pm$0.0655
& (99.99722$\pm$1.2)$\times10^{-3}$
& (2.78$\pm$1.20)$\times10^{-3}$
& 52.63$\pm$12.40
& 47.37$\pm$12.40 \\

Fixed + No Band
& 17.35$\pm$2.00
& 0.7560$\pm$0.1541
& 0.2635$\pm$0.1542
& (99.99779$\pm$1.0)$\times10^{-3}$
& (2.21$\pm$1.00)$\times10^{-3}$
& 72.02$\pm$9.10
& 27.98$\pm$9.10 \\

\hline
\end{tabular*}
\end{table*}
As shown in Table~\ref{tab:fusion_band_ablation}, $\mathcal{L}_{\mathrm{band}}$ increases the KAN high-frequency proportion and improves reconstruction quality under both fusion strategies; under additive fusion, it also further suppresses the already small MLP high-frequency response. Fixed routing yields a substantially larger KAN high-frequency
proportion but markedly lower reconstruction accuracy due to the restricted coefficient assignment, indicating that hard frequency assignment is overly restrictive. Additive fusion with band-separation regularization therefore provides a better balance between frequency-oriented coordination and reconstruction flexibility.

\subsection{Parameter-Controlled Comparison}
To investigate whether the performance improvement mainly comes from the increased model capacity, we conduct an additional parameter-controlled experiment. Specifically, the compared methods are adjusted toward a parameter scale comparable to that of our model (4.13M), subject to their architecture-specific parameterization constraints. Detailed network configurations are provided in the supplementary material.

The results are summarized in Table~\ref{tab:parameter_matched_comparison}. Under this parameter-controlled setting, our method still achieves strong reconstruction performance, indicating that the observed improvement cannot be explained solely by a larger trainable-parameter budget.

\begin{table}[!t]
\centering
\caption{Parameter-controlled comparison of image fitting performance on the DIV2K dataset. The compared methods are adjusted toward a trainable-parameter scale comparable to that of the proposed method (4.13M), subject to their architecture-specific parameterization constraints. Results are reported as mean $\pm$ standard deviation.}
\label{tab:parameter_matched_comparison}

\scriptsize
\renewcommand{\arraystretch}{1.5}
\setlength{\tabcolsep}{0pt}

\begin{tabular*}{\columnwidth}
{@{\extracolsep{\fill}}lccc@{}}
\toprule

\textbf{Method}
& \textbf{PSNR (dB)} $\uparrow$
& \textbf{SSIM} $\uparrow$
& \textbf{LPIPS} $\downarrow$ \\
\midrule

FINER
& $41.65 \pm 2.746$
& $0.9853 \pm 0.0090$
& $0.0234 \pm 0.0207$ \\

MFN
& $30.19 \pm 3.027$
& $0.9193 \pm 0.0278$
& $0.1531 \pm 0.0402$ \\

FR
& $41.66 \pm 3.848$
& $0.9726 \pm 0.0320$
& $0.0477 \pm 0.0559$ \\

MLP-PE
& $29.24 \pm 1.821$
& $0.8085 \pm 0.0381$
& $0.2885 \pm 0.0626$ \\

SIREN
& $40.74 \pm 2.662$
& $0.9910 \pm 0.0024$
& $0.0209 \pm 0.0111$ \\

WIRE
& $39.22 \pm 6.887$
& $0.9585 \pm 0.0666$
& $0.0443 \pm 0.0474$ \\

KAN
& $28.36 \pm 3.163$
& $0.8791 \pm 0.0530$
& $0.2336 \pm 0.0698$ \\

BACON
& $39.32 \pm 2.146$
& $0.9803 \pm 0.0086$
& $0.0510 \pm 0.0370$ \\

\textbf{Ours}
& $\mathbf{45.72 \pm 2.349}$
& $\mathbf{0.9912 \pm 0.0030}$
& $\mathbf{0.0030 \pm 0.0020}$ \\

\bottomrule
\end{tabular*}

\vspace{0.8mm}
\begin{minipage}{\columnwidth}
\footnotesize
\end{minipage}

\end{table}

\section{Discussion}
The experiments on 1D square-wave and audio signals, 2D images, 3D volumes, SDFs, videos, and 4D LFs demonstrate the applicability of the proposed framework to signals with different dimensionalities and frequency characteristics. The MLP-KAN architecture provides two heterogeneous representation pathways, while the wavelet module exposes and coordinates their frequency responses through DWT, additive coefficient fusion, and band-separation regularization. The resulting representation is frequency-oriented: the MLP branch exhibits a stronger low-frequency tendency, while the KAN branch provides relatively stronger responses to localized variations and detail components. Both branches remain involved in the reconstruction because the corresponding wavelet coefficients are fused additively.

The ablation experiments further clarify the roles of the fusion strategy and spectral regularization. Removing the band-separation loss consistently reduces reconstruction quality across the evaluated 1D, 2D, 3D, and 4D tasks. The branch-wise wavelet-energy analysis also shows that the regularization increases the relative high-frequency response of the KAN branch and further suppresses the small high-frequency response of the MLP branch. The KAN branch still contains substantial low-frequency energy, indicating a frequency-oriented tendency rather than complete spectral separation. Fixed routing produces a stronger concentration of KAN energy in the detail bands, but leads to substantially lower reconstruction accuracy. These observations indicate that hard frequency assignment is too restrictive for the current dual-branch representation, while additive fusion with band-separation regularization provides a better balance between frequency coordination and reconstruction flexibility. The parameter-controlled comparison further shows that the reconstruction advantage cannot be attributed solely to a larger parameter budget, since the proposed method maintains strong performance when the compared methods are adjusted toward a parameter scale comparable to that of the proposed model.

The wavelet decomposition level, KAN grid size, spline degree, and branch architectures remain dependent on the signal dimensionality and data characteristics. The framework also involves a trade-off between representation accuracy, model complexity, and computational efficiency. Future work will investigate lighter heterogeneous branch designs, adaptive frequency-domain coordination, and more efficient CUDA implementations to improve the accuracy--efficiency balance while preserving reconstruction fidelity.

\section{Conclusion}
We present a spatial--frequency-aware implicit neural representation framework that integrates MLP and KAN branches through wavelet-domain decomposition and additive fusion, together with band-separation regularization. The proposed design exploits the different frequency tendencies of the two branches while allowing both to contribute to the reconstructed signal. The spectral regularization further promotes complementary branch responses without imposing exclusive subband assignment.

Experiments on 1D signals, 2D images, 3D volumes and SDFs, videos, and 4D LFs demonstrate its effectiveness across diverse signal dimensions and modalities. Controlled ablations verify the contribution of the band separation regularization and show that additive fusion provides higher reconstruction fidelity than fixed routing under the evaluated settings. Parameter-controlled experiments further indicate that the performance improvement is not explained solely by model size. These results support wavelet-domain coordination of heterogeneous neural representations as an effective approach for multidimensional implicit signal fitting.

\section*{Acknowledgment}
The authors thank doctoral researchers Dingbo Hou and Yulin Han and engineers Yao Guo and Lin Nai from the Institute of Computational Imaging, Beijing Information Science and Technology University, for their assistance with the comparative experiments. Specifically, Dingbo Hou was mainly responsible for the light-field fitting experiments, Yulin Han for the video fitting experiments, Yao Guo for the SDF and voxel fitting experiments, and Lin Nai for the image fitting experiments. The parameter ablation experiments were completed jointly by the aforementioned four individuals and the authors, and the remaining experiments not mentioned above were completed by the authors.

\bibliographystyle{IEEEtran}  
\bibliography{ref}            
\vfill
\end{document}